\documentclass[a4paper,fleqn]{cas-dc}

\usepackage[numbers]{natbib}
\usepackage{amsmath}
\usepackage{graphicx}
\usepackage{caption}
\usepackage{subcaption}
\usepackage{booktabs}
\usepackage{xcolor}
\usepackage{algorithm}
\usepackage{algorithmic}
\usepackage{todonotes}
\usepackage{makecell}

\definecolor{linkblue}{HTML}{1181AA}

\hypersetup{
    colorlinks=true,
    linkcolor=linkblue,
    citecolor=linkblue,
    urlcolor=linkblue
}

\RenewDocumentCommand{\printorcid}{}{}

\newcommand{\figref}[1]{\hyperref[#1]{\textcolor{blue}{Figure~\ref*{#1}}}}
\renewcommand{\tabref}[1]{\hyperref[#1]{\textcolor{blue}{Table~\ref*{#1}}}}

\newcommand{\eqnref}[1]{Eq.~\hyperref[#1]{\textcolor{blue}{(\ref*{#1})}}}
\newcommand{\secref}[1]{Section~\hyperref[#1]{\textcolor{blue}{\ref*{#1}}}}
\newcommand{\subsecref}[1]{Subsection~\hyperref[#1]{\textcolor{blue}{\ref*{#1}}}}

\newcommand{\algref}[1]{\hyperref[#1]{\textcolor{blue}{Algorithm~\ref*{#1}}}}

\def\tsc#1{\csdef{#1}{\textsc{\lowercase{#1}}\xspace}}
\tsc{WGM}
\tsc{QE}

\begin{document}
\let\WriteBookmarks\relax
\def\floatpagepagefraction{1}
\def\textpagefraction{.001}

\shorttitle{Symmetry Matters: Auditing and Symmetrizing 3D Generative Models}
\shortauthors{N. Caytuiro and I. Sipiran}

\title [mode = title]{Symmetry Matters: Auditing and Symmetrizing 3D Generative Models}  



%

\author[1]{Nicolas Caytuiro}

\cormark[1]


\ead{ncaytuir@dcc.uchile.cl}


\credit{Writing – original draft, Resources, Visualization, Methodology, Investigation, Data curation, Conceptualization}

\author[1]{Ivan Sipiran}

\ead{isipiran@dcc.uchile.cl}


\credit{Writing – review \& editing, Supervision, Validation, Resources, Project administration, Conceptualization}

\cortext[1]{Corresponding author}

\affiliation[1]{organization={University of Chile},
           addressline={Ave. Alameda Libertador Bernardo O'Higgins 1058}, 
           city={Santiago},
           postcode={8330111},
           country={Chile}}


\begin{abstract}
Symmetry is a strong prior present in many object categories, yet standard benchmarks for 3D generative models rarely report whether this prior is preserved. We study symmetry preservation in unconditional point cloud generation. We first audit the symmetry of generated shapes by several 3D generative models and compute a normalized symmetry score based on the Chamfer Distance (CD). We show that although current 3D generative models achieve competitive results under standard evaluation, they reveal a persistent symmetry gap when a symmetry-aware evaluation protocol is applied. To test whether this gap is merely inherited from the training data, we evaluate these models over a mirrored-objects dataset derived from ShapeNet and analyze symmetry dynamics during training. Mechanism-inspired diagnostic tests were conducted at the sampling and latent-representation levels to further show that reflection symmetry is not reliably encoded in the learned generative process. Finally, to address this gap, we propose a data-centric symmetry-based intervention: training generative models on a half-objects dataset and reconstructing full objects by reflection during sampling. Across multiple backbones, this intervention substantially improves geometric consistency and visual plausibility while remaining competitive under standard metrics. These findings suggest that symmetry-aware evaluation is needed alongside standard benchmarks, and future 3D generative models should incorporate this prior explicitly, either during training or sampling. Project page and code: \url{https://shapevision.dcc.uchile.cl/symmetry-matters/}
\end{abstract}




\begin{keywords}
Mechanism-Inspired Diagnostics \sep 3D Generative Models \sep Reflection Symmetry \sep Geometry-Aware Evaluation \sep Point Cloud Generation
\end{keywords}

\maketitle

\begin{figure*} 
    \centering
    \includegraphics[width=\linewidth]{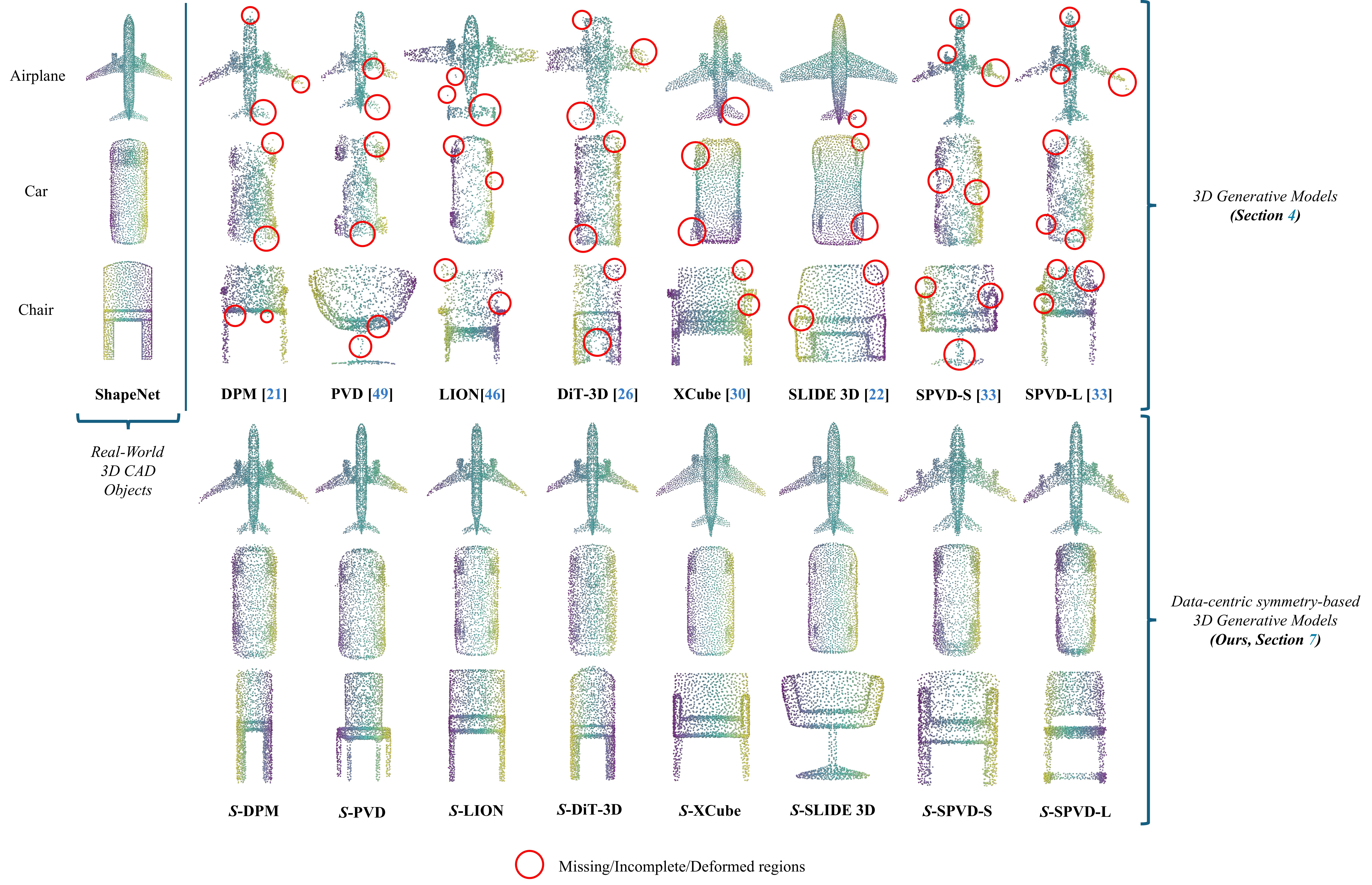}
    \caption{\emph{Top}: Unconditional shape generation with 2,048 points for each class-specific model. \emph{Bottom}: Results obtained with our data-centric symmetry-based intervention (\protect\secref{sec:proposal}).
    Red circles highlight missing, incomplete, and deformed regions.}
    \label{fig:generated_shapes}
\end{figure*}

\section{Introduction}

Generative models are often treated as black boxes. In text, image, audio, and 3D generation alike, one typically provides a prompt, a latent code, random noise, or partial views, and then evaluates the resulting sample almost exclusively at the output level. This black-box view has been sufficient for rapid progress in generative modeling, but it obscures a central scientific question: \emph{what internal computations allow these models to preserve—or destroy—structural properties already present in the training data?} In 3D generation, this question is especially important because geometric priors such as symmetry, continuity, and part consistency are not cosmetic attributes; they are fundamental properties of shape organization \cite{mitra_Symmetry3DGeometry_2013, treder_LookingGlassReviewHuman_2010, xu_SurveyDeepLearningbased_2023}.

Symmetry is one of the most pervasive of these priors; it is ubiquitous in both natural and artificial objects, and it is a fundamental property in geometry. It serves as a crucial visual cue that helps humans perceive object structure and interpret spatial relationships \cite{treder_LookingGlassReviewHuman_2010, li_SymmetryStrikesBack_2024}. In computer vision, either in the 2D or 3D domain, symmetry has long been leveraged as a structural constraint to simplify complex visual tasks such as pose estimation \cite{merrill_SymmetryUncertaintyAwareObject_2022a, zhao_LearningSymmetryAwareGeometry_2023}, shape synthesis \cite{li_LearningPartGeneration_2020b, levy_SymmCDSymmetryPreservingCrystal_2024}, and 3D reconstruction \cite{rock_Completing3DObject_2015, wu_UnsupervisedLearningProbably_2020, xu_LadybirdQuasiMonteCarlo_2020}. Motivated by these developments, various methods have been proposed to collect evidence of symmetric correspondences under a set of symmetry transformations and to detect symmetry from completed or well-observed 3D shapes where sufficient evidence exists \cite{Sinha_detectingandreconstructing_2012, mitra_Symmetry3DGeometry_2013, zhou_NeRDNeural3D_2021, Shi_learningtodetect_2023, je_RobustSymmetryDetection_2024, yang_SYM3DLearningSymmetric_2024, li_SymmetryStrikesBack_2024}.

While current 3D generative models exhibit strong performance in synthesizing shapes from 3D or depth data \cite{zhou_3DShapeGeneration_2021b, zeng_LIONLatentPoint_2022, ren_XCubeLargeScale3D_2024, lyu_SLIDEControllableMesh_2023, ren_TIGERTimeVaryingDenoising_2024}, explicitly generating symmetrical objects is rarely an architectural or experimental design objective. This remains true even though many objects in widely used 3D shape datasets exhibit clear reflection symmetry with respect to the canonical plane $x=0$. As illustrated in \figref{fig:generated_shapes}, objects from the ShapeNet dataset \cite{chang_ShapeNetInformationRich3D_2015} highlight this property. Recent approaches face this challenge by incorporating symmetry into their architecture \cite{levy_SymmCDSymmetryPreservingCrystal_2024, lu_StructurePreservingDiffusion_2025, yang_SYM3DLearningSymmetric_2024, kelvinius_WyckoffDiffGenerativeDiffusion_2025}, but they still lack problem generalization. 

In this context, a central research question arises: \emph{do current 3D generative models preserve the symmetry present in the training data?} We address this question by conducting an in-depth and systematic study of symmetry preservation across several representative 3D generative models. First, an initial audit reveals a consistent symmetry gap between real and generated samples. To determine whether the observed symmetry gap is simply inherited from the data, we construct a controlled mirrored-objects dataset in which the reflection symmetry prior is made more explicit by design. We then retrain the representative models on this new distribution and compare their results against the original references using a symmetry measurement protocol. The resulting analysis shows that stronger symmetry in the training distribution does not automatically translate into symmetry-aware generation. Through a symmetry dynamics study during training, we show that symmetry failure is not merely a static data artifact but part of the learned generative process itself.

Second, we draw inspiration from mechanistic interpretability to design diagnostic tests for 3D generative models in a concrete task, that is, assessing whether the sampling process and latent representations behave consistently under reflection. Mechanistic interpretability aims to explain model behavior by identifying the internal computations and representations that implement it, rather than relying on post-hoc output inspection \cite{lin_SurveyMechanisticInterpretability_2025, somvanshi_BridgingBlackBox_2026}. While this perspective has been developed extensively for language models, recent work has begun extending it to multimodal systems and, more recently, to 3D diffusion transformers \cite{plattner_CircuitsDynamicsUnderstanding_2026}. However, we do not claim to identify internal circuits, causal components, activation pathways, or representation features responsible for symmetry breaking. Instead, our goal is to provide mechanism-inspired evidence about where reflection inconsistency becomes visible in the generative pipeline. 

Finally, we show that a simple data-centric intervention can recover a large portion of the lost structure. Instead of modifying the architecture or loss function, we train representative models on half-objects and reconstruct full shapes by reflection. Our results show that the resulting models produce substantially more symmetric, geometrically consistent, and visually plausible objects. At the same time, the generated shapes achieve competitive results against those of the original models across standard metrics such as Coverage and the 1-Nearest Neighbor accuracy (1-NNA), along with CD and EMD, which measure distributional similarity but do not directly capture high-level structural priors such as symmetry.

The contributions of this paper are as follows:

\begin{itemize}
    \item {We introduce a systematic baseline for analyzing reflection symmetry in 3D shapes, enabling a quantitative assessment of symmetry preservation in both real-world datasets and generated shapes. This addresses an important dimension of shape quality that is largely overlooked in current benchmarks.}
    \item {Through an extensive empirical study of representative state-of-the-art 3D generative models, we show that symmetry is not consistently preserved in generated shapes, despite being strongly present in the training data. Our results reveal a clear structural gap between real and generated samples from baselines, even when traditional metrics indicate competitive generative performance.}
    \item {We introduce mechanism-inspired diagnostic tests for studying symmetry preservation in 3D generative models. These tests examine reflection consistency at the sampling and latent-representation levels, providing complementary evidence beyond output-space metrics.}
    \item {We propose a data-centric symmetry-based intervention based on half-objects training that substantially improves reflection symmetry across several representative 3D generative models.}
\end{itemize}

\section{Related Works}\label{sec:related_works}

\subsection{Symmetry}

``\emph{Symmetry is what we see at a glance.}'' -- Blaise Pascal.

According to \citeauthor{je_RobustSymmetryDetection_2024}~\cite{je_RobustSymmetryDetection_2024}, symmetry plays a crucial role in Computer Vision and Computer Graphics, and is widely used in various downstream tasks, including 3D reconstruction \cite{chen_AutoSweepRecovering3D_2020, xu_LadybirdQuasiMonteCarlo_2020, allingham_GenerativeModelSymmetry_2024}, shape refinement \cite{mitra_Symmetrization_2007}, image manipulation \cite{zhang_PortraitShadowManipulation_2020}; and is extended to different 3D shape representations such as 3D point clouds \cite{ji_FastEfficient3D_2019, sun_SymmetryDetectionAnalysis_2021}, meshes \cite{allingham_GenerativeModelSymmetry_2024}, voxel grids \cite{tulsiani_ObjectCentricMultiViewAggregation_2020}, and implicit fields \cite{xu_LadybirdQuasiMonteCarlo_2020, je_RobustSymmetryDetection_2024}. These priors enable more robust shape processing and help regularize ill-posed tasks, specifically under noisy or partial observations.

Despite its prevalence, symmetry is often treated as a post-processing step (e.g., symmetrization \cite{mitra_Symmetrization_2007}) or used implicitly in loss functions without directly constraining the generative process. As a result, many deep learning models are not explicitly symmetry-aware during training or inference, leading to asymmetrical or geometrically inconsistent outputs when data is sparse or noisy. This limitation has motivated a growing interest in embedding symmetry more directly into the architecture and training of 3D generative models—explored further in \secref{sec:symmetry_aware_3d_gen_models}.

\subsection{3D Reflection Symmetry}

Reflectional symmetry is the most fundamental form of symmetry observed. Symmetries among shape parts can also be discovered via transform-space voting, assuming that the symmetric parts only differ by a \emph{rigid transformation} \cite{xu_PartialIntrinsicReflectional_2009}. Symmetrical objects are widespread in nature, architecture, and art, and reflectional symmetry plays a crucial role in human visual perception \cite{mitra_Symmetry3DGeometry_2013}. This concept is being widely used in 3D computer vision research. A significant research effort in this domain focuses on applying reflectional symmetry principles to enhance the synthesis of 3D shapes.

As noted by \citeauthor{yang_SYM3DLearningSymmetric_2024}~\cite{yang_SYM3DLearningSymmetric_2024}, symmetric objects can be split into two halves, where one half approximately mirrors the other. Based on this principle, we train generative models using one half of each object and reconstruct full shapes by reflecting the generated outputs across the symmetry plane.

\subsection{Symmetry-aware 3D Generative Models} \label{sec:symmetry_aware_3d_gen_models}

Current 3D generative frameworks, including GANs, VAEs, and, more recently, diffusion models, rely on data-driven learning to implicitly capture symmetrical patterns. However, this strategy can be sample-inefficient, especially when the training data exhibit diverse or spatial symmetries.

Recent research has begun to address this limitation by explicitly incorporating symmetry into either the model architecture or the training objective. For instance, \citeauthor{wu_UnsupervisedLearningProbably_2020}~~\cite{wu_UnsupervisedLearningProbably_2020} propose an unsupervised method for learning 3D object categories from raw single-view images based on the idea that many object categories have, at least in principle, a symmetric structure. NeRD \cite{zhou_NeRDNeural3D_2021} focuses on detecting 3D reflection symmetry by estimating mirror planes, while GET3D \cite{gao_GET3DGenerativeModel_2022} generates textured meshes with rich geometry through dual-branch generation. \citeauthor{niu_SymmetryawareAlignmentMethod_2023}~\cite{niu_SymmetryawareAlignmentMethod_2023} introduce a symmetry-aware alignment pipeline that incrementally refines object orientation based on symmetry cues, such as features, planes, symmetry axes, and frontal orientations. \citeauthor{allingham_GenerativeModelSymmetry_2024}~\cite{allingham_GenerativeModelSymmetry_2024} directly model approximate symmetries within generative pipelines. Additionally, \citeauthor{je_RobustSymmetryDetection_2024}~\cite{je_RobustSymmetryDetection_2024} propose a robust symmetry detection framework that leverages Langevin Dynamics in a refined symmetry space. However, these methods are not directly applied to 3D generation. 

More recent work has extended these ideas to diffusion-based generative frameworks. For instance, SymmCD \cite{levy_SymmCDSymmetryPreservingCrystal_2024} encodes crystallographic symmetries within the generative process, while WyckoffDiff \cite{kelvinius_WyckoffDiffGenerativeDiffusion_2025} introduces symmetry-aware Wyckoff representations. Following these advances, \citeauthor{lu_StructurePreservingDiffusion_2025}~\cite{lu_StructurePreservingDiffusion_2025} introduce \emph{structure-preserving diffusion processes}, a family of diffusion processes for learning distributions that possess additional structure, such as group symmetries, by developing theoretical conditions under which the diffusion transition steps preserve said symmetry. Despite their relevance, these methods are still tailored to highly structured domains, particularly crystallographic and molecular generation, where the relevant symmetries are explicitly defined by physical or algebraic constraints. 

While these works focus on incorporating symmetry through architectural or training constraints, they do not explicitly analyze how symmetry is preserved under standard training setups. This leaves the symmetry behavior in existing generative pipelines largely unexplored, motivating a complementary analysis-based perspective. Our work complements symmetry-aware generation by auditing the behavior of state-of-the-art 3D generative models before proposing an architecture-agnostic remedy.

\subsection{Mechanism-Inspired Diagnostics for 3D Generative Models}

Mechanistic Interpretability aims to identify the internal components and computations that implement model behavior, rather than explaining outputs solely through feature attribution or post-hoc visualization. Research in this field emphasizes circuit discovery, causal interventions, sparse feature analysis, and activation-level exploration as central tools for relating observed behavior to internal mechanisms~\cite{somvanshi_BridgingBlackBox_2026}.

This perspective is beginning to extend from language models to multimodal systems, but recent literature still emphasizes a substantial gap between what is known for language-only models and what is understood for cross-modal or generative systems \cite{lin_SurveyMechanisticInterpretability_2025}. The recent 3D-diffusion case study of meltdown in diffusion transformers is particularly relevant: activation patching localizes the failure to a specific cross-attention activation and connects those internal states to a symmetry-breaking bifurcation in the reverse process \cite{plattner_CircuitsDynamicsUnderstanding_2026}. 

However, our work does not attempt to provide this level of mechanistic explanation. In particular, we do not perform activation patching, causal ablations, sparse feature decomposition, or circuit tracing. Instead, we use mechanistic interpretability as an inspiration for designing diagnostic tests--at the sampling and latent-representation levels--that complement output-level inspection.

\section{Background and Preliminaries}\label{sec:preliminaries} 

\subsection{Reflection Symmetry}

Reflectional symmetry is a common trait in many object categories \cite{mitra_Symmetry3DGeometry_2013, yang_SYM3DLearningSymmetric_2024}.

Let $S \subset \mathbb{R}^3$ denote a set of homogeneous surface points of a 3D object. The object is reflectionally symmetric with respect to a plane $p$ if reflecting every point $x \in S$ across $p$ produces another point that also belongs to $S$. Following prior works, this can be expressed as $S = M_p(S)$, where $M_p$ is the reflection operator associated with plane $p$ and surface properties are preserved under reflection \cite{zhou_NeRDNeural3D_2021,li_SymmetryStrikesBack_2024}.

In practice, many ShapeNet object-categories are approximately symmetric. Our goal is therefore not to enforce exact mathematical symmetry on every sample, but to quantify the degree to which generated samples preserve the bilateral regularity present in the training distribution.

\subsection{Householder Transformation}

For a symmetry plane with normal vector $\mathbf{n}$ and a point $\mathbf{m}$ on the plane, reflection can be expressed as a generalized Householder transformation. The matrix $\mathcal{A} = \mathbf{I} - 2\mathbf{n}\mathbf{n}^\top$ defines reflection about the corresponding plane through the origin, and an offset term $\mathbf{t} = 2\mathbf{n}\mathbf{n}^\top \mathbf{m}$ extends the transformation to arbitrary planes. Any point $\mathbf{p}$ is then reflected as $\mathbf{R}(\mathbf{p})=\mathcal{A}\mathbf{p} + \mathbf{t}$ \cite{rock_Completing3DObject_2015}.

We use this technique to mirror an object over the YZ plane, which is orthogonal to the $X$-axis (i.e., the plane defined by $x=0$).

\subsection{Geometric and Distributional Metrics}

\subsubsection{{Chamfer Distance}} 
The Chamfer Distance (CD) is a standard metric for evaluating the geometric fidelity between a generated shape and a ground-truth shape. It is used extensively in 3D reconstruction, shape completion, and point set generation, as well as in higher-level metrics such as 1-NNA, MMD, and Coverage, which rely on pairwise distances. Formally, the CD between two point clouds $S_1$ and $S_2 \subseteq \mathbb{R}^3$ is defined as \cite{fan_point_2016, aguirre_DatasetfreeApproachSelfsupervised_2025}:

\begin{equation}\label{eq:chamfer_distance}
\begin{split}
CD(S_1, S_2) = & \frac{1}{|S_1|} \sum_{x \in S_1} \min_{y \in S_2} \|x - y\|_2^2 \\
& + \frac{1}{|S_2|} \sum_{y \in S_2} \min_{x \in S_1} \|x - y\|_2^2
\end{split}
\end{equation}

The lower the CD value, the better.

\subsubsection{{Earth Mover's Distance}} 
The Earth Mover's Distance (EMD) measures the optimal cost of transforming one point cloud into another. It solves an optimization problem, namely, \emph{the assignment problem} \cite{achlioptas_learning_2018}. For all but a zero-measure subset of point set pairs, an optimal bijection $\phi$ is unique and invariant under infinitesimal movement of the points~\cite{fan_point_2016}. Although EMD is computationally more expensive than CD, it provides a more accurate and globally consistent measure of geometric similarity.


\subsubsection{Coverage (COV)}

Proposed by \citeauthor{achlioptas_learning_2018}~\cite{achlioptas_learning_2018}, COV measures the fraction of real samples that have a generated sample within some distance threshold (or that are the nearest neighbor match). It reflects how well the generated set covers the models of the target distribution. Higher is better.

COV has become standard for assessing diversity vs. fidelity in 3D point cloud generation.

\subsubsection{1-Nearest Neighbor Accuracy (1-NNA)}

Introduced by \citeauthor{lopez-paz_revisiting_2017}~\cite{lopez-paz_revisiting_2017} as a classifier-based two-sample test. The 1-NN accuracy (1-NNA) pools real and generated samples and uses a 1-NN classifier (leave-one-out) to identify each sample's source; the accuracy of this test indicates if the two distributions are identical (50\% is ideal for identical distributions). Applied later by ~\cite{achlioptas_learning_2018} to evaluate 3D generative models. The 1-NN accuracy (denoted 1-NNA) directly measures the distributional similarity between generated and real 3D point sets.

\subsection{Symmetry Measurement Protocol}\label{sec:symmetry_measurement_protocol}

To evaluate the degree of reflection symmetry in a 3D shape, we propose a quantitative analysis based on point cloud representations. Specifically, for each object in a given class, we represent its 3D shape as a point cloud $\mathcal{P} = \{\mathbf{x}_i \in \mathbb{R}^3 \ |\ i = 1, \dots, N\}$, and compute its reflected counterpart $\mathcal{P}^\prime$ using the Householder transformation $\mathbf{M}$ with normal vector $\mathbf{n} = [1, 0, 0]^\top$. This reflection is defined with respect to the plane $x = 0$. The resulting reflected point cloud is then defined by $\mathcal{P}^\prime = \{\mathbf{x}_i^\prime = \mathbf{M}\mathbf{x}_i \ | \ \mathbf{x}_i \in \mathcal{P}\}$.


To quantify the degree of symmetry for aligned objects with respect to the canonical symmetry plane $x=0$, we propose a Normalized Symmetry score based on CD (NSCD), defined as follows:

\begin{equation}
\text{NSCD}(\mathcal{P}) = \frac{CD(\mathcal{P}, \mathcal{R}(\mathcal{P}))}{diag(\mathcal{P})^2}
\end{equation}

We compute the CD (defined in \eqnref{eq:chamfer_distance}) between the original point cloud $\mathcal{P}$ and its reflected counterpart $\mathcal{R}(\mathcal{P}) = \mathcal{P}^\prime$, and then normalize the resulting value using the diagonal $diag(\mathcal{P})^2$ of the object's axis-aligned bounding box (AABB). We normalize the CD by the squared diagonal length of the object bounding box to remove scale dependence. Since CD is based on squared Euclidean distances, larger objects would otherwise exhibit larger symmetry errors even under comparable relative geometric deviations.

Following prior work on reflection symmetry and symmetry-aware alignment~\cite{podolak_PlanarreflectiveSymmetryTransform_2006, mitra_Symmetry3DGeometry_2013, niu_SymmetryawareAlignmentMethod_2023, zhou_NeRDNeural3D_2021}, object categories whose dominant symmetry planes differ from $x=0$ could, in principle, be canonicalized or rigidly aligned before computing NSCD.

\section{Symmetry Audit of Current 3D Generative Models}\label{sec:symmetry_audit} 


\subsection{Symmetry Prior in ShapeNet}\label{sec:symmetry_current_datasets}

In the context of 3D shape generation, data plays a crucial role in guiding both the learning and synthesis processes. Using an appropriate dataset can help the learning and generation processes based on the application~\cite{xu_SurveyDeepLearningbased_2023}.

In this work, we use the ShapeNet dataset \cite{chang_ShapeNetInformationRich3D_2015} as a reference to assess the presence of symmetry in real-world 3D objects. ShapeNet is a large-scale dataset containing richly annotated 3D CAD models across a wide variety of categories. Following previous works on point cloud generation \cite{yang_PointFlow3DPoint_2019, zhou_3DShapeGeneration_2021b, luo_DiffusionProbabilisticModels_2021, zeng_LIONLatentPoint_2022, mo_DiT3DExploringPlain_2023, ren_XCubeLargeScale3D_2024, ren_TIGERTimeVaryingDenoising_2024}, we focus on three representative and commonly used classes: \emph{Airplane}, \emph{Car}, and \emph{Chair}, which exhibit diverse symmetrical objects.

\paragraph{\textbf{Symmetry Distribution.}}\label{sec:symmetry_dist_original} 

\begin{figure}
    \centering
    \includegraphics[width=1\columnwidth]{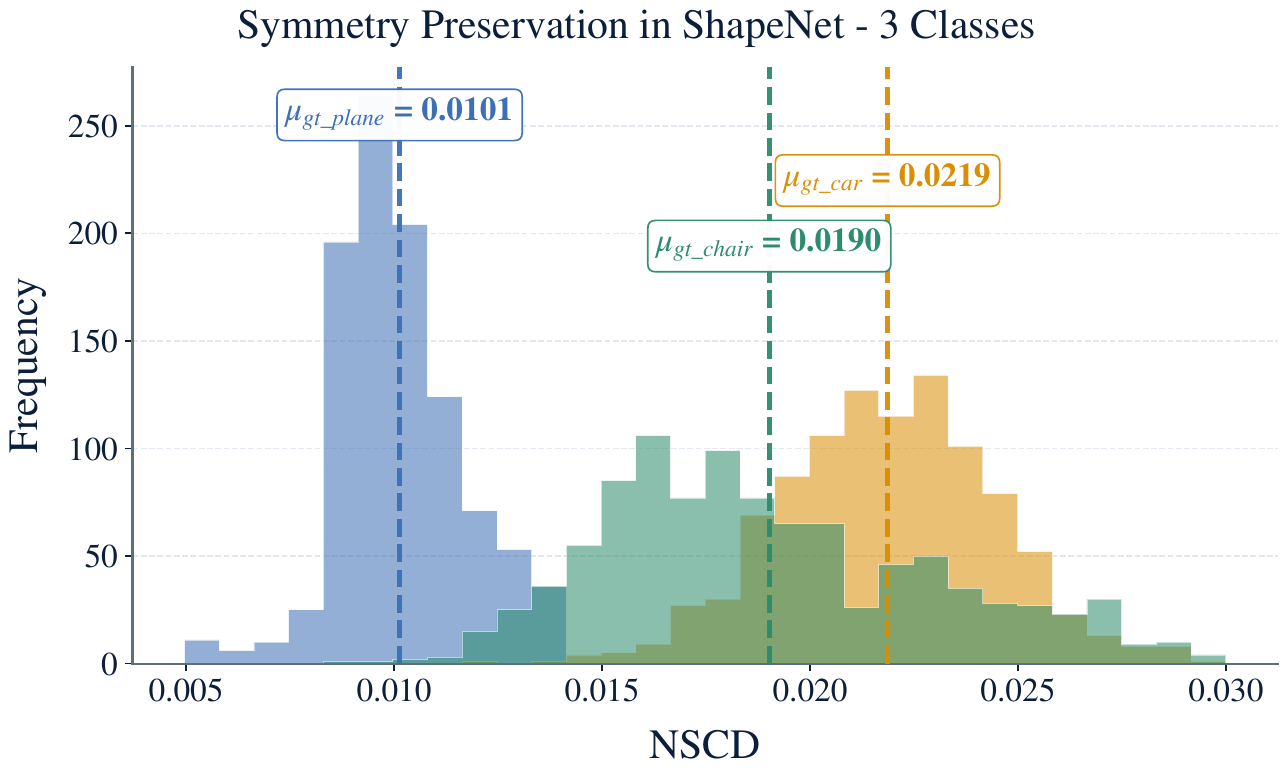}
    \caption{Reflection symmetry in ShapeNet.}
    \label{fig:shapenet_symmetry_histogram}
\end{figure}

We apply the symmetry measurement protocol described in \secref{sec:symmetry_measurement_protocol} to 1,000 training samples from the selected ShapeNet classes. Lower mean NSCD values indicate better symmetry.

As shown in \figref{fig:shapenet_symmetry_histogram}, the mean NSCD values are \emph{0.0101} for Airplane, \emph{0.0219} for Car, and \emph{0.0190} for Chair. Also, the chair and car classes show a broader symmetry distribution, indicating greater intra-class variability; we further analyze this effect in subsequent sections.

The first observation is crucial: symmetry is not a marginal phenomenon in ShapeNet. It is a prominent structural regularity already present in the training distribution. Any systematic asymmetry in generated samples should therefore be interpreted not as an arbitrary stylistic choice, but as a failure to preserve a strong prior encoded in the data.

\subsection{Symmetry Gap in Generated Samples}\label{sec:symmetry_current_3dgenmodels}

\begin{figure*}[h]
    \centering
    \begin{subfigure}{0.325\textwidth}
        \centering
        \includegraphics[width=\linewidth]{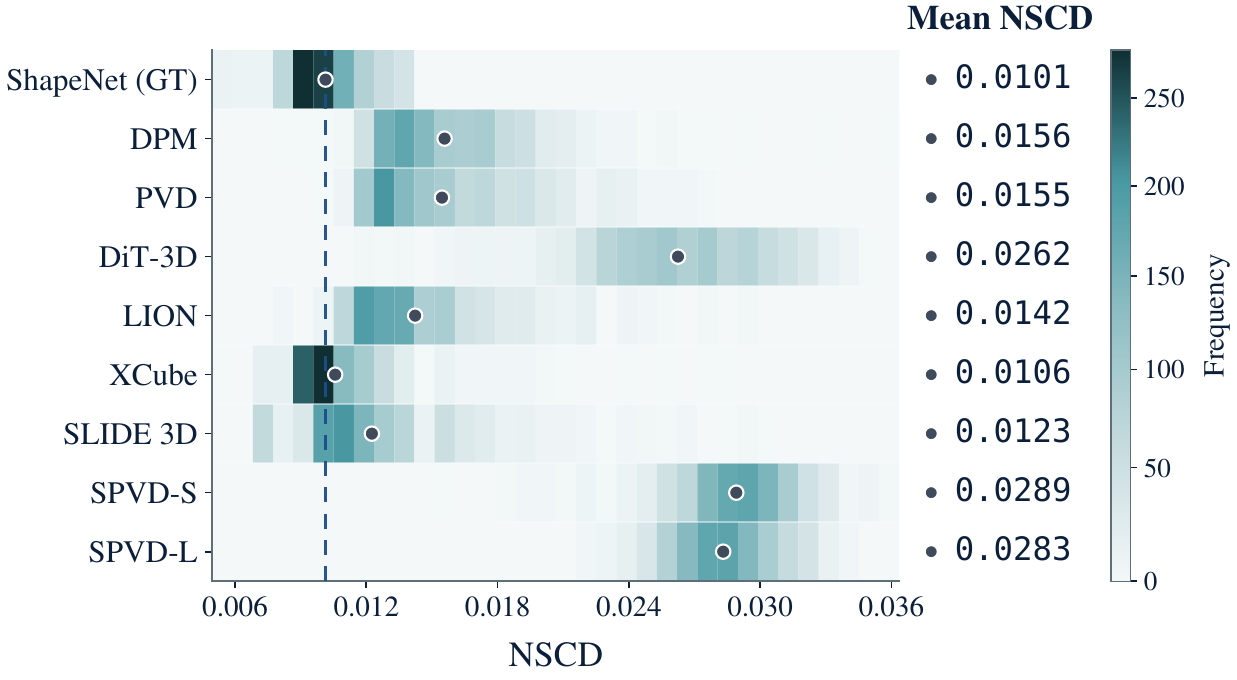}
        \caption{Airplane class}
        \label{fig:hist_airplane}
    \end{subfigure}
    \hfill
    \begin{subfigure}{0.325\textwidth}
        \centering
        \includegraphics[width=\linewidth]{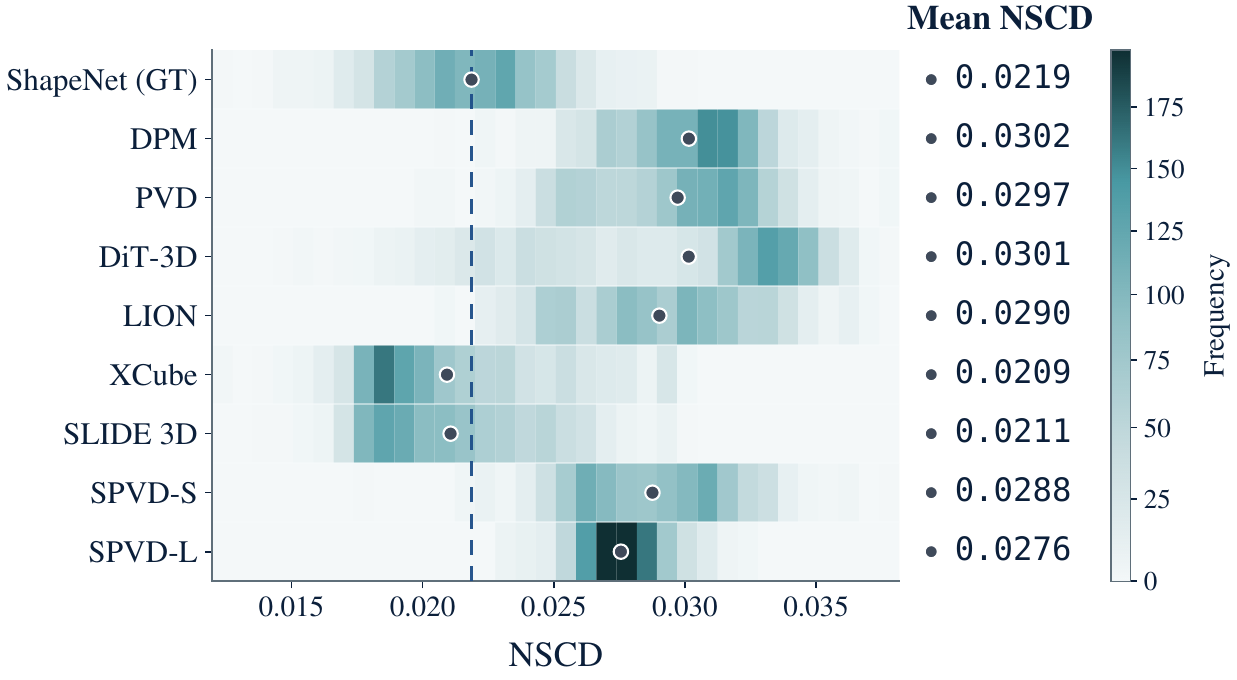}
        \caption{Car class}
        \label{fig:hist_car}
    \end{subfigure}
    \hfill
    \begin{subfigure}{0.325\textwidth}
        \centering
        \includegraphics[width=\linewidth]{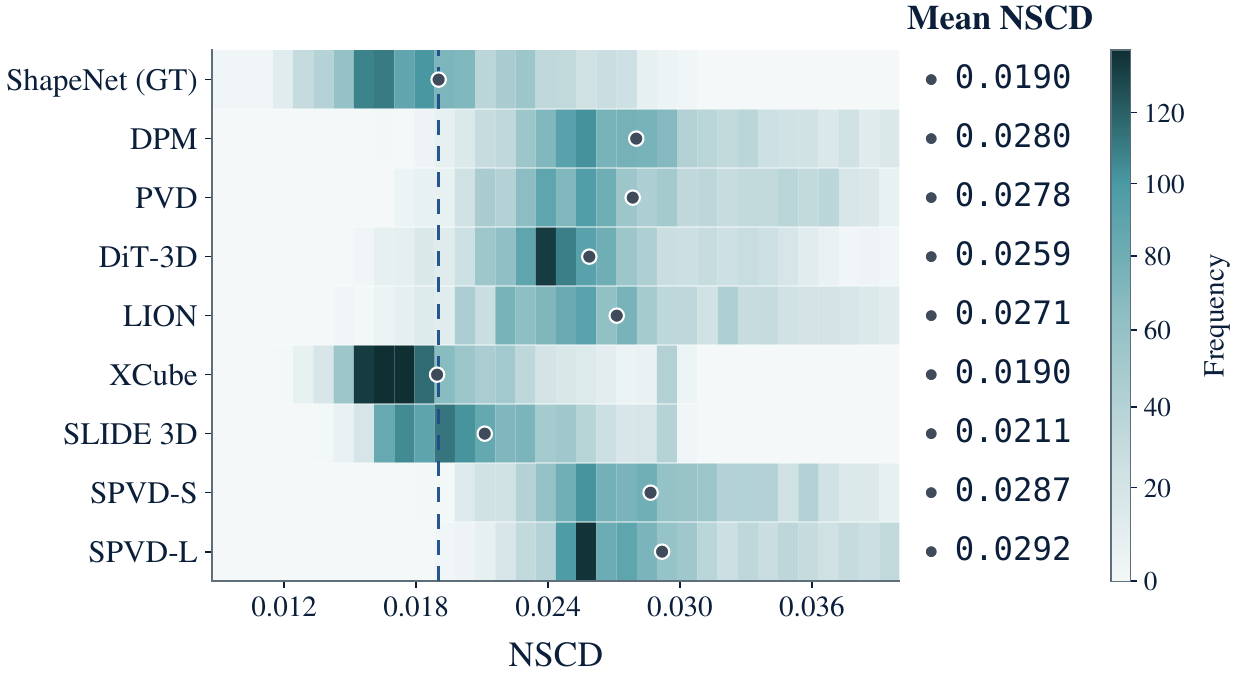}
        \caption{Chair class}
        \label{fig:hist_chair}
    \end{subfigure}
    
    \caption{\textbf{Symmetry evaluation across 3D generative models}. Results correspond to the models trained over the \textbf{original ShapeNet dataset}. Ground-truth results from ShapeNet are provided for reference. The vertical dashed line highlights the mean symmetry score of the ground-truth distribution.}
    \label{fig:original_models_histograms}
\end{figure*}

As discussed in \secref{sec:related_works}, various deep learning approaches have been proposed for 3D shape synthesis. In this work, we focus on evaluating eight representative models: DPM \cite{luo_DiffusionProbabilisticModels_2021}, PVD \cite{zhou_3DShapeGeneration_2021b}, LION \cite{zeng_LIONLatentPoint_2022}, DiT-3D \cite{mo_DiT3DExploringPlain_2023}, XCube \cite{ren_XCubeLargeScale3D_2024} (with a patch size of 4), SLIDE 3D \cite{lyu_SLIDEControllableMesh_2023}, and SPVD \cite{romanelis_EfficientScalablePoint_2025} (`S' and `L' variants). All these models were trained on the ShapeNet dataset in the three classes.

For our experiments, we perform inference using the publicly available pre-trained checkpoints. For all models, we sample 1,000 objects with 2,048 points. For models that do not natively generate this resolution, such as XCube, we apply Farthest Point Sampling (FPS) \cite{li_AdjustableFarthestPoint_2022} to ensure an equal number of points.
Subsequently, we apply the symmetry measurement protocol to the generated samples. 

As shown in \figref{fig:original_models_histograms}, across all tested models, the symmetry distribution shifts away from the ground-truth ShapeNet distribution. PVD and LION already exhibit larger average symmetry errors than the data across all three classes, while DiT-3D, SPVD-S, and SPVD-L show even larger deviations in several classes. XCube and SLIDE 3D are more competitive across all three classes, although they still exhibit non-negligible deviations from the ground-truth distribution. We think that this behavior does not necessarily imply that these models explicitly learn reflection symmetry. Rather, it may result from architectural or training biases that favor smoother, more regular, or more prototypical shapes, thereby suppressing small asymmetric details present in the data. In the next section, we further examine this possibility by constructing a more explicitly symmetrized version of ShapeNet and evaluating whether training 3D generative models on this controlled distribution improves symmetry preservation.

Moreover, as shown in \figref{fig:generated_shapes}, the generated shapes are neither visually plausible nor geometrically symmetric, revealing missing and outlier points, deformed regions, and sparse point coverage, further highlighting the model's limitations in maintaining structural and geometric consistency, such as symmetry.

This output-space audit provides the first layer of evidence for a symmetry gap. However, output-space statistics alone cannot determine whether the gap is caused by the training distribution, by sampling noise, or by internal properties of the learned generative map. The next sections address those possibilities.

\section{Symmetry Prior Verification with a Controlled Mirrored Dataset}\label{sec:symmetry_controlled_dataset}

A natural objection is that the observed symmetry gap may be a dataset artifact. Perhaps the original ShapeNet distribution is only approximately symmetric, and the models are merely reproducing this approximation. To test this, we build a new controlled mirrored-objects dataset derived from the original ShapeNet by taking each object $X \in \{airplane, \, car, \, chair\}$ and concatenating it with its reflected counterpart $\mathcal{R}(X)$ by applying the Householder transformation and then resampling them back to 15,000 points using FPS. This process creates an explicitly symmetrized reference distribution.

\begin{figure}
    \centering
    \includegraphics[width=1\columnwidth]{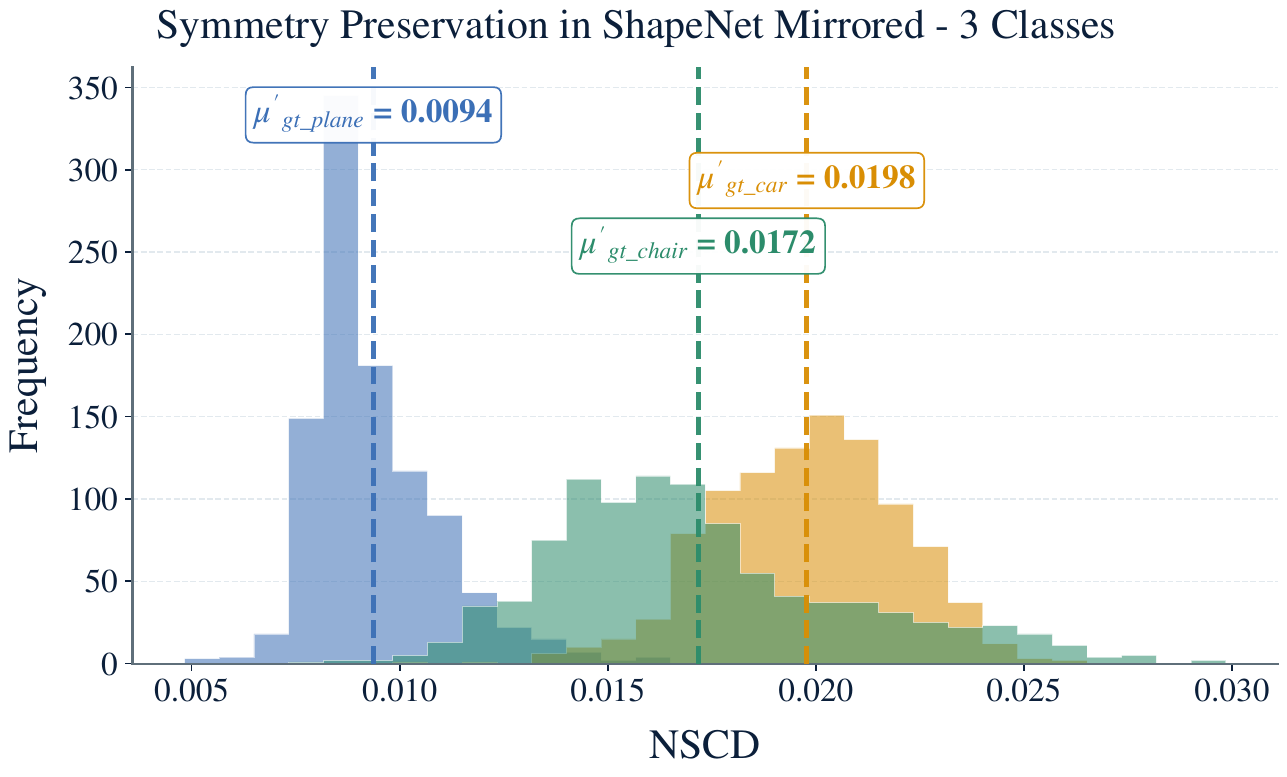}
    \caption{Reflection symmetry in mirrored ShapeNet objects.}
    \label{fig:shapenet_mirrored_symmetry_histogram}
\end{figure}

We then repeat the same symmetry distribution analysis described in \secref{sec:symmetry_dist_original}. As shown in \figref{fig:shapenet_mirrored_symmetry_histogram}, the mirrored objects exhibit substantially lower NSCD values, while preserving almost the same class-level distributional behavior. This experiment shows that ShapeNet objects can be made more symmetric through mirroring without substantially altering the overall structure of the data distribution.

\begin{table}[width=1\linewidth,cols=7,pos=ht]
  \centering
  \caption{Mean symmetry scores ($\mu_{NSCD}$) for generated samples. Mean NSCD values are the same as reported in \protect\figref{fig:original_models_histograms} for the original models. We further report standard deviation values ($\pm$) for quantitative comparisons. Mean and standard deviation values are multiplied by $1 \times 10^3$.}
  \label{tab:symmetry-preservation-results-1}
  

    \resizebox{\linewidth}{!}{%
    \begin{tabular}{@{}lcccccc@{}}
        \toprule
        \multirow{2.5}{*}{\textbf{Model}} & \multicolumn{2}{c}{Airplane} & \multicolumn{2}{c}{Car} & \multicolumn{2}{c}{Chair} \\
        \cmidrule(lr){2-3} \cmidrule(lr){4-5} \cmidrule(lr){6-7}
        & Orig. & Mirr. & Orig. & Mirr. & Orig. & Mirr. \\
        \midrule
        DPM       & 15.6 $\pm$ 2.8 & 15.8 $\pm$ 2.8 & 30.2 $\pm$ 2.3 & 31.1 $\pm$ 11.2 & 28.0 $\pm$ 4.6 & 28.9 $\pm$ 5.4 \\
        PVD       & 15.5 $\pm$ 3.2 & 15.5 $\pm$ 3.3 & 29.7 $\pm$ 2.9 & 29.6 $\pm$ 2.9  & 27.8 $\pm$ 5.1 & 27.5 $\pm$ 5.2 \\
        DiT-3D    & 26.2 $\pm$ 3.7 & 24.7 $\pm$ 3.0 & 30.1 $\pm$ 4.7 & 21.8 $\pm$ 2.7  & 25.9 $\pm$ 4.3 & 17.6 $\pm$ 1.5 \\
        LION      & 14.2 $\pm$ 3.0 & 15.1 $\pm$ 3.3 & 29.0 $\pm$ 3.1 & 29.1 $\pm$ 3.1  & 27.1 $\pm$ 5.2 & 26.9 $\pm$ 5.5 \\
        XCube     & 10.6 $\pm$ 2.0 & 15.2 $\pm$ 1.6 & 20.9 $\pm$ 3.2 & 25.7 $\pm$ 2.3  & 19.0 $\pm$ 3.8 & 25.7 $\pm$ 3.9 \\
        SLIDE 3D  & 12.3 $\pm$ 3.4 & 119.0 $\pm$ 24.0 & 21.1 $\pm$ 2.8 & 49.5 $\pm$ 12.8 & 21.1 $\pm$ 3.6 & 62.4 $\pm$ 19.6 \\
        SPVD-S    & 28.9 $\pm$ 2.4 & 18.8 $\pm$ 3.8 & 28.8 $\pm$ 2.7 & 31.5 $\pm$ 2.4  & 28.7 $\pm$ 4.5 & 30.1 $\pm$ 4.7 \\
        SPVD-L    & 28.3 $\pm$ 2.1 & 15.8 $\pm$ 2.9 & 27.6 $\pm$ 1.4 & 31.0 $\pm$ 2.6  & 29.2 $\pm$ 4.7 & 32.3 $\pm$ 5.7 \\
        \bottomrule
    \end{tabular}
    }

    \vspace{0.2em}
    \noindent
    \begin{minipage}{\linewidth}
    \raggedright
    \tiny
    \textbf{Orig.}: Original model.\\[-0.4ex]
    \textbf{Mirr.}: Model trained over the mirrored-objects dataset.
    \end{minipage}
\end{table}

\begin{table}[width=1\linewidth,cols=7,pos=ht]
  \centering
  \caption{\textbf{Symmetry preservation results.} Absolute error $\Delta$ between the samples' average symmetry and the ground truth distribution mean ($\mu_{gt}$ and $\mu'_{gt}$), where $\Delta_1 = |\mu_{gt} - \mu_{NSCD\_orig}|$ and $\Delta_2 = |\mu'_{gt} - \mu_{NSCD\_mirr}|$. Bold values indicate that those models learn from the dataset's symmetric distribution and decrease the absolute deviation ($\Delta$). We further report standard deviation values ($\pm$) for quantitative comparisons. Mean and standard deviation values are multiplied by $1 \times 10^3$.}
  \label{tab:symmetry-preservation-results}


    \resizebox{\linewidth}{!}{%
    \begin{tabular}{@{}lcccccc@{}}
        \toprule
        \multirow{2.5}{*}{\textbf{Model}} & \multicolumn{2}{c}{Airplane} & \multicolumn{2}{c}{Car} & \multicolumn{2}{c}{Chair} \\
        \cmidrule(lr){2-3} \cmidrule(lr){4-5} \cmidrule(lr){6-7}
        & $\Delta_1$ & $\Delta_2$ & $\Delta_1$ & $\Delta_2$ & $\Delta_1$ & $\Delta_2$ \\
        \midrule
        DPM      & 5.4 $\pm$ 0.1 & 6.4 $\pm$ 0.1 & 8.3 $\pm$ 0.1 & 11.3 $\pm$ 0.4 & 9.0 $\pm$ 0.2 & 11.7 $\pm$ 0.2 \\
        PVD      & 5.3 $\pm$ 0.1 & 6.1 $\pm$ 0.1 & 7.9 $\pm$ 0.1 & 9.8 $\pm$ 0.1 & 8.8 $\pm$ 0.2 & 10.4 $\pm$ 0.2 \\
        DiT-3D & 16.1 $\pm$ 0.1 & 15.4 $\pm$ 0.1 & 8.3 $\pm$ 0.2 & \textbf{2.0 $\pm$ 0.1} & 6.9 $\pm$ 0.2 & \textbf{0.4 $\pm$ 0.1} \\
        LION     & 4.1 $\pm$ 0.1 & 5.8 $\pm$ 0.1 & 7.2 $\pm$ 0.1 & 9.3 $\pm$ 0.1 & 8.1 $\pm$ 0.2 & 9.7 $\pm$ 0.2 \\
        XCube    & 0.4 $\pm$ 0.1 & 5.9 $\pm$ 0.1 & 0.9 $\pm$ 0.1 & 5.9 $\pm$ 0.1 & 0.1 $\pm$ 0.2 & 7.0 $\pm$ 0.2 \\
        SLIDE~3D & 2.1 $\pm$ 0.1 & 109.6 $\pm$ 0.8 & 0.8 $\pm$ 0.1 & 29.8 $\pm$ 0.4 & 2.1 $\pm$ 0.2 & 45.3 $\pm$ 0.6 \\
        SPVD-S & 18.8 $\pm$ 0.1 & \textbf{9.4 $\pm$ 0.1} & 6.9 $\pm$ 0.1 & 11.7 $\pm$ 0.1 & 9.6 $\pm$ 0.2 & 12.9 $\pm$ 0.2 \\ 
        SPVD-L & 18.2 $\pm$ 0.1 & \textbf{6.4 $\pm$ 0.1} & 5.7 $\pm$ 0.1 & 11.2 $\pm$ 0.1 & 10.2 $\pm$ 0.2 & \textbf{15.1 $\pm$ 0.2} \\
        \midrule
        \textbf{Constraint} & \multicolumn{2}{c}{$\Delta_2 \geq \Delta_1$} & \multicolumn{2}{c}{$\Delta_2 \geq \Delta_1$} & \multicolumn{2}{c}{$\Delta_2 \geq \Delta_1$} \\
        \bottomrule
    \end{tabular}
    }

    \vspace{0.2em}
    \noindent
    \begin{minipage}{\linewidth}
    \raggedright
    \tiny
    \textbf{$\Delta_1$}: Original model.\\[-0.4ex]
    \textbf{$\Delta_2$}: Model trained over the mirrored-objects dataset.
    \end{minipage}
\end{table}

Using the new mirrored-objects dataset, we retrain all the models tested in the previous section. For a set of $N = 1,000$ generated samples, we report the mean symmetry score $\mu_{NSCD} = \frac{1}{N} \sum_{i=1}^{N} NSCD(X, \mathcal{R}(X))$, as reported in \tabref{tab:symmetry-preservation-results-1}, where $X$ is the original point cloud, and $\mathcal{R}(X)$ is its reflected counterpart. We also compare the mean values against the original ShapeNet mean $\mu_{gt}$ and the mirrored-objects dataset mean $\mu'_{gt}$ through two absolute deviations $\Delta_1$ and $\Delta_2$.

As shown in \tabref{tab:symmetry-preservation-results}, if models fully learn and preserve the symmetry prior of the mirrored dataset, one would expect $\Delta_2$ to decrease sharply, ideally toward zero or less than $\Delta_1$. Instead, the current results show that for all models, the three classes still satisfy $\Delta_2 > \Delta_1$, with some exceptions. In other words, even when the training distribution is made explicitly more symmetric, the generated samples remain farther from the new symmetric reference than the original-model outputs were from the original distribution. This suggests that the observed symmetry gap is not only inherited from the data; it is also induced by the learned generative process. In the next section, we investigate it further.

\paragraph{\textbf{Symmetry Dynamics During Training.}}

\begin{figure*}[h]
    \centering
    \begin{subfigure}{0.325\textwidth}
        \centering
        \includegraphics[width=\linewidth]{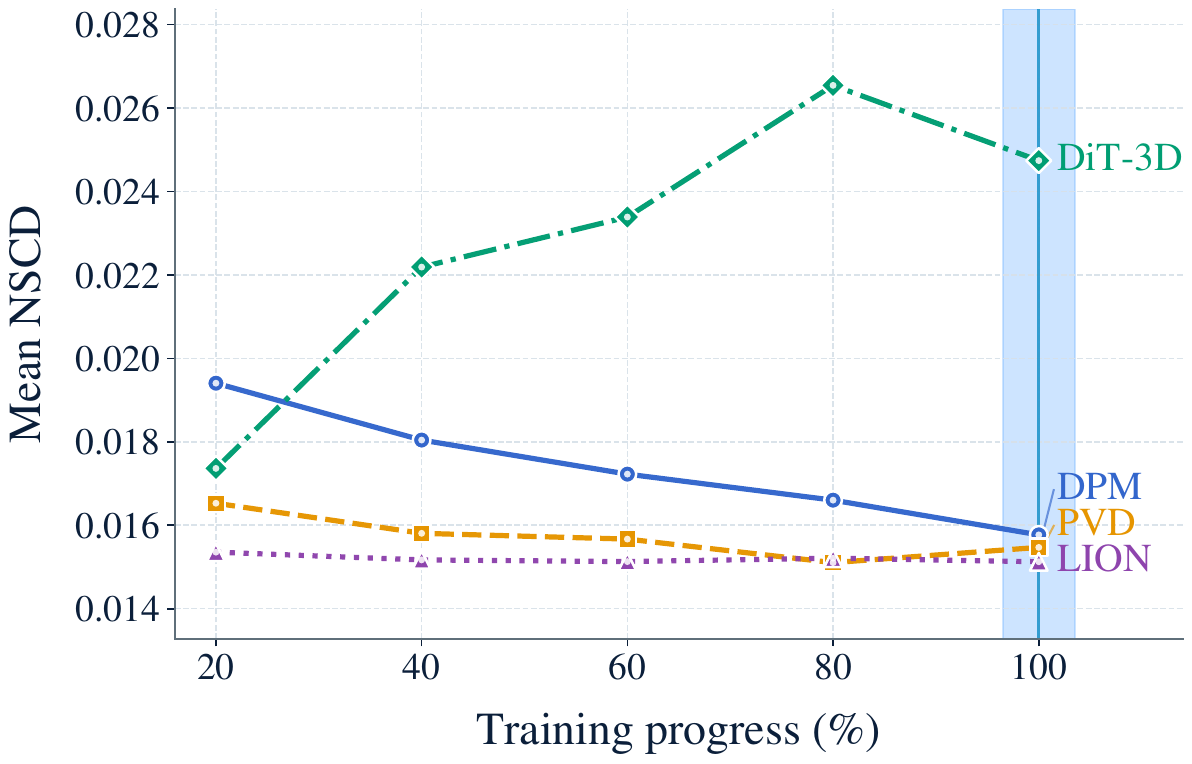}
        \caption{Airplane class}
        \label{fig:per-epoch_airplane}
    \end{subfigure}
    \hfill
    \begin{subfigure}{0.325\textwidth}
        \centering
        \includegraphics[width=\linewidth]{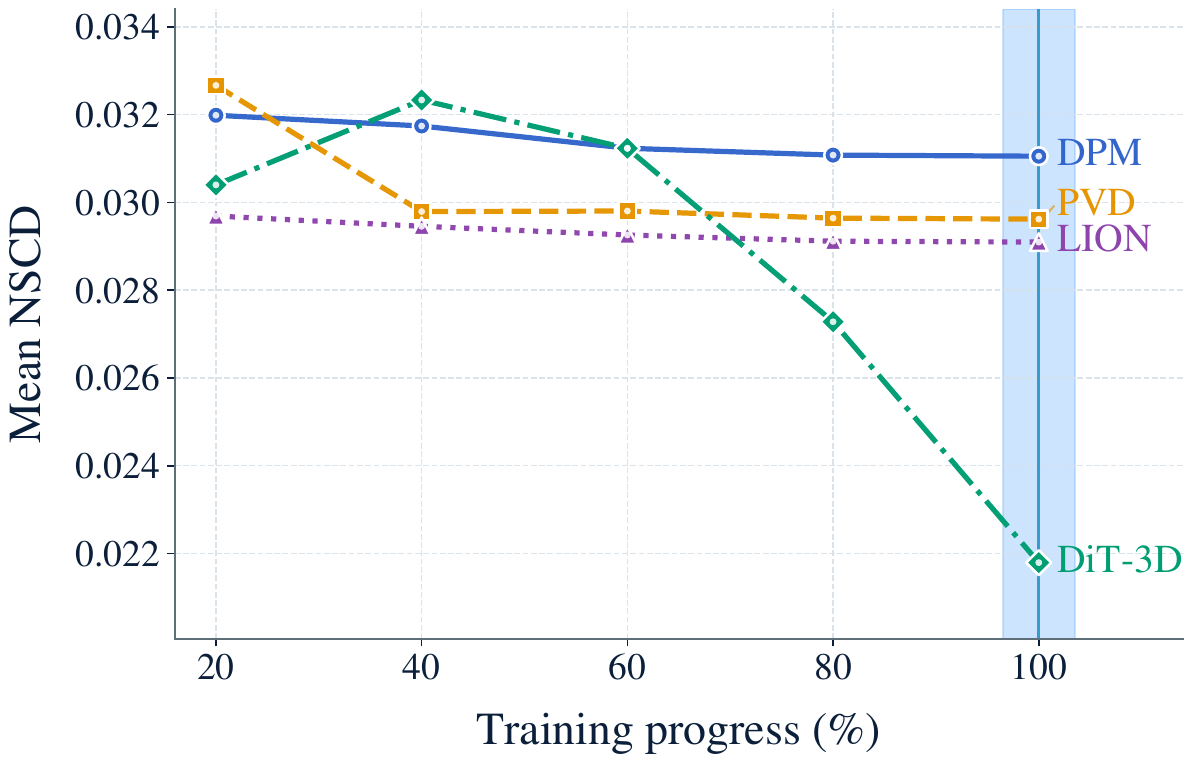}
        \caption{Car class}
        \label{fig:per-epoch_car}
    \end{subfigure}
    \hfill
    \begin{subfigure}{0.325\textwidth}
        \centering
        \includegraphics[width=\linewidth]{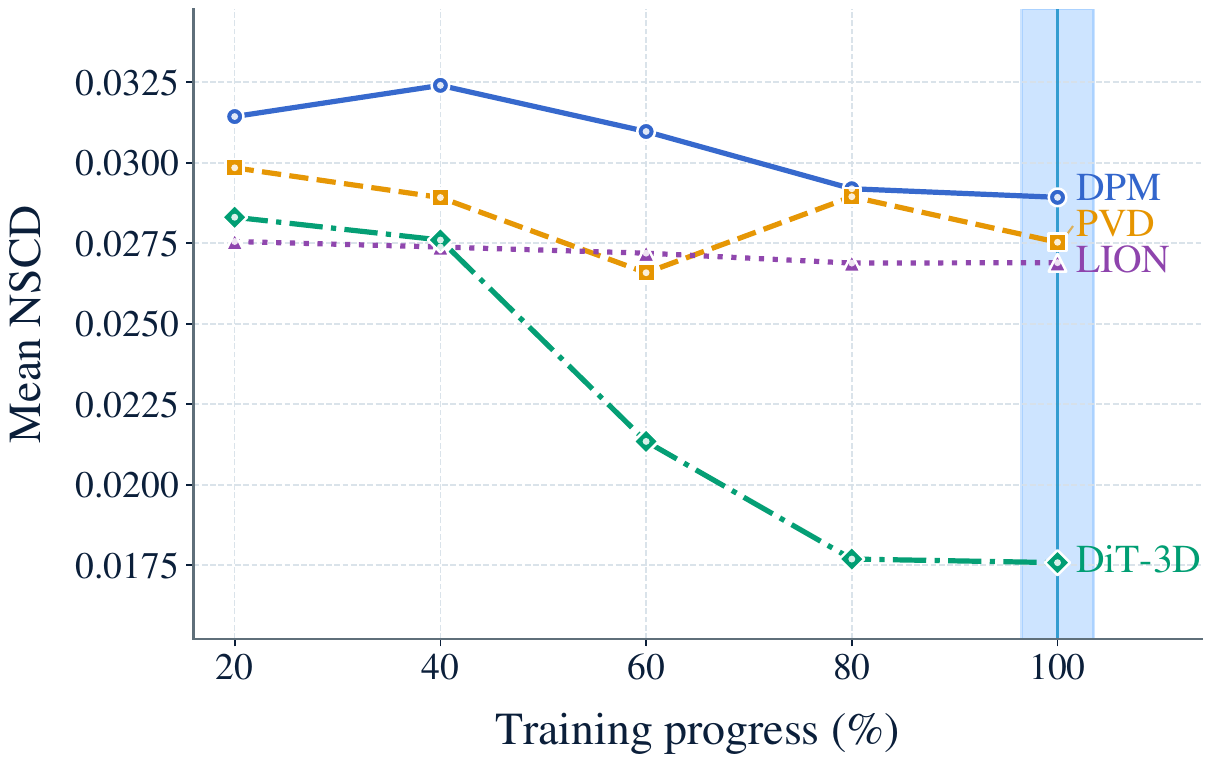}
        \caption{Chair class}
        \label{fig:per-epoch_chair}
    \end{subfigure}
    
    \caption{\textbf{Evolution of symmetry during training.} Symmetry is evaluated every 20\% of the training process across different 3D generative models and object categories.}
    \label{fig:per-epoch_analysis}
\end{figure*}

\figref{fig:per-epoch_analysis} shows the evolution of symmetry throughout the training process for each model and object category. For this analysis, we consider four representative 3D generative models covering different generative paradigms: DPM, a diffusion-based model for point cloud generation; PVD, which combines point and voxel representations within a diffusion framework; LION, which integrates a VAE with latent diffusion; and DiT-3D, a diffusion-transformer-based generative model. Symmetry is evaluated at five checkpoints corresponding to 20\%, 40\%, 60\%, 80\%, and 100\% of training. At each checkpoint, a set of $N=1{,}000$ samples is generated, and the mean symmetry score $\mu_{\text{NSCD}}$ is computed.

Across all categories, we observe that symmetry improves progressively as training advances. While models such as DPM, PVD, and LION exhibit relatively stable or gradually improving symmetry in the first two classes (e.g., airplane and car), more complex models, such as DiT-3D, show higher variability throughout the optimization process. While it achieves competitive symmetry at early training stages in the airplane class, its performance can degrade as training progresses. This suggests that symmetry is not consistently preserved during optimization, and such instability highlights that higher model capacity alone does not necessarily guarantee better structural consistency.


Overall, these results suggest that the symmetry gap is not merely inherited from the data distribution but is actively shaped by the learning dynamics of the generative models. 

\section{Mechanism-Inspired Diagnostics of Symmetry Breaking}

The previous section shows that symmetry degradation persists even when the data are explicitly symmetrized. We next ask where this degradation enters the pipeline. Rather than relying only on output-space evaluation (as presented in \secref{sec:symmetry_audit}), we adopt a mechanism-inspired diagnostic perspective; targeted counterfactual probes are used to test whether internal computations behave consistently under a symmetry transformation \cite{somvanshi_BridgingBlackBox_2026, lin_SurveyMechanisticInterpretability_2025, plattner_CircuitsDynamicsUnderstanding_2026}. We ask two concrete questions: (1) \emph{If a diffusion model were symmetry-aware, would reflecting the initial Gaussian noise produce a reflected final shape?}, and (2) \emph{If a latent diffusion model encoded structural and geometrical properties coherently, would a shape and its mirrored version map to nearby latent representations?}

\subsection{Diffusion Sampling Inspection}

For all models that generate samples by applying the diffusion process \cite{ho_DenoisingDiffusionProbabilistic_2020}, we assess symmetry consistency directly in the sampling (backward) process. Starting from a standard Gaussian prior $p(\mathbf{x}_T)$, where $T$ is the number of steps in the backward process, we generate a second input by reflecting the same noise across the $x = 0$ symmetry plane $\mathcal{R}(p(\mathbf{x}_T))$. If the denoising process were symmetry-aware, both trajectories should produce reflected versions of the same object. We sampled $N = 1,000$ objects per class, computed the CD for each pair of objects, and averaged the results. We can formalize this process as follows:

\begin{equation}
\mu_{cd} = \frac{1}{N} \sum_{i=1}^{N} NSCD(\mathcal{G}({\mathbf{x}_T}_i), \, \mathcal{G}(\mathcal{R}({\mathbf{x}_T}_i))),
\end{equation}

where $\mathcal{G}$ is the generative model that transforms an initial Gaussian prior (noise) $\mathbf{x}_T \sim p(\mathbf{x}_T)$ into a generated sample $\mathbf{x}_0 \sim q(\mathbf{x}_0)$.

We repeat the same process for the models trained on the mirrored-objects dataset obtained in \secref{sec:symmetry_controlled_dataset}. 

\begin{table}[width=1\linewidth,cols=7,pos=ht]
  \centering
  \caption{Quantitative comparison of the symmetry-breaking inspection. Bold values highlight cases where training on the mirrored-objects dataset reduces the symmetry-breaking score with respect to the original models. We further report standard deviation values ($\pm$) for quantitative comparisons. Mean and standard deviation values are multiplied by $1 \times 10^4$.}

    \resizebox{\linewidth}{!}{%
    \begin{tabular}{@{}lcccccc@{}}
    \toprule
    \multirow{2.5}{*}{\textbf{Model}} & \multicolumn{2}{c}{Airplane} & \multicolumn{2}{c}{Car} & \multicolumn{2}{c}{Chair} \\
    \cmidrule(lr){2-3} \cmidrule(lr){4-5} \cmidrule(lr){6-7}
    & Orig. & Mirr. & Orig. & Mirr. & Orig. & Mirr. \\
    \midrule
    DPM      & 4.91 $\pm$ 0.96 & 5.09 $\pm$ 1.01 & 7.21 $\pm$ 1.18 & 7.38 $\pm$ 0.84 & 8.41 $\pm$ 2.33 & 8.56 $\pm$ 2.30 \\
    PVD      & 6.13 $\pm$ 2.72 & 6.18 $\pm$ 2.80 & 6.17 $\pm$ 2.21 & \textbf{6.10 $\pm$ 2.17} & 14.6 $\pm$ 5.70 & \textbf{14.2 $\pm$ 5.73} \\
    DiT-3D   & 11.7 $\pm$ 8.09 & 13.0 $\pm$ 12.3 & 17.9 $\pm$ 24.8 & 61.8 $\pm$ 22.1 & 13.6 $\pm$ 5.38 & 21.0 $\pm$ 40.2 \\
    XCube    & 8.45 $\pm$ 4.57 & \textbf{5.28 $\pm$ 3.10} & 6.36 $\pm$ 2.33 & \textbf{6.02 $\pm$ 2.99} & 20.9 $\pm$ 8.36 & \textbf{12.6 $\pm$ 4.80} \\
    SLIDE~3D & 0.42 $\pm$ 0.12 & 4.36 $\pm$ 1.19 & 1.02 $\pm$ 0.24 & 3.26 $\pm$ 0.55 & 1.41 $\pm$ 0.52 & 10.1 $\pm$ 3.24 \\
    SPVD-S   & 2.99 $\pm$ 3.70 & 6.72 $\pm$ 3.33 & 6.24 $\pm$ 2.45 & \textbf{5.26 $\pm$ 2.01} & 14.3 $\pm$ 5.84 & \textbf{12.7 $\pm$ 5.78} \\
    SPVD-L   & 5.20 $\pm$ 4.51 & 5.42 $\pm$ 2.75 & 4.60 $\pm$ 1.30 & 5.99 $\pm$ 1.79 & 16.2 $\pm$ 15.3 & \textbf{15.00 $\pm$ 7.35} \\
    \midrule
    \midrule
    LION~(E) & 12.09 $\pm$ 6.09 & \textbf{6.87 $\pm$ 2.64} & 7.68 $\pm$ 4.89 & \textbf{5.01 $\pm$ 1.90} & 8.61 $\pm$ 6.07 & \textbf{4.38 $\pm$ 1.93} \\
    \bottomrule
    \end{tabular}
    }

    \vspace{0.2em}
    \noindent
    \begin{minipage}{\linewidth}
    \raggedright
    \tiny
    \textbf{Orig.}: Original model. \\[-0.4ex]
    \textbf{Mirr.}: Model trained over the mirrored-objects dataset. \\[-0.4ex]
    \textbf{Orig. ($\mathbf{E}$)}: Original encoder. \\[-0.4ex]
    \textbf{Mirr. ($\mathbf{E}$)}: Encoder trained over the mirrored-objects dataset.
    \end{minipage}
  \label{tab:symmetry-breaking-inspection}
\end{table}

As shown in \tabref{tab:symmetry-breaking-inspection}, the average values are not closer to zero for any class, and the mirrored-objects dataset training does not eliminate it. The models may learn a distribution containing many symmetric objects, but they do not learn a denoising process whose trajectories commute with reflection. These findings contradict the noise properties stated by \cite{liang_SymmetryAllYou_2025}: (1) Symmetry allows for more accurate predictions of $\mathbf{x}_0$ from $\mathbf{x}_T$. (2) Symmetric noise models can be inverted effectively during denoising. (3) Symmetry ensures that information is not skewed or lost in certain directions.

We also observe that the Car and Chair classes tend to exhibit higher average values. As stated in \secref{sec:symmetry_dist_original} and \secref{sec:symmetry_controlled_dataset}, these two classes present broader symmetry distributions than Airplane, which may partially explain the deviations. Moreover, several models show higher average values after training on the mirrored-objects dataset, suggesting that explicitly symmetrizing the training data does not necessarily lead to symmetry-preserving training or generation. The reported standard deviations provide complementary evidence; large values, as observed for DiT-3D in Car and Chair and for SPVD-L in Chair, indicate that the symmetry-breaking scores vary substantially across generated samples, suggesting outlier generations, severe geometric artifacts, or unstable sampling trajectories.


\subsection{Latent Representation Inspection}

\begin{figure*}[t]
  \centering
  \begin{minipage}{\linewidth}
    \centering
    \includegraphics[width=1\linewidth]{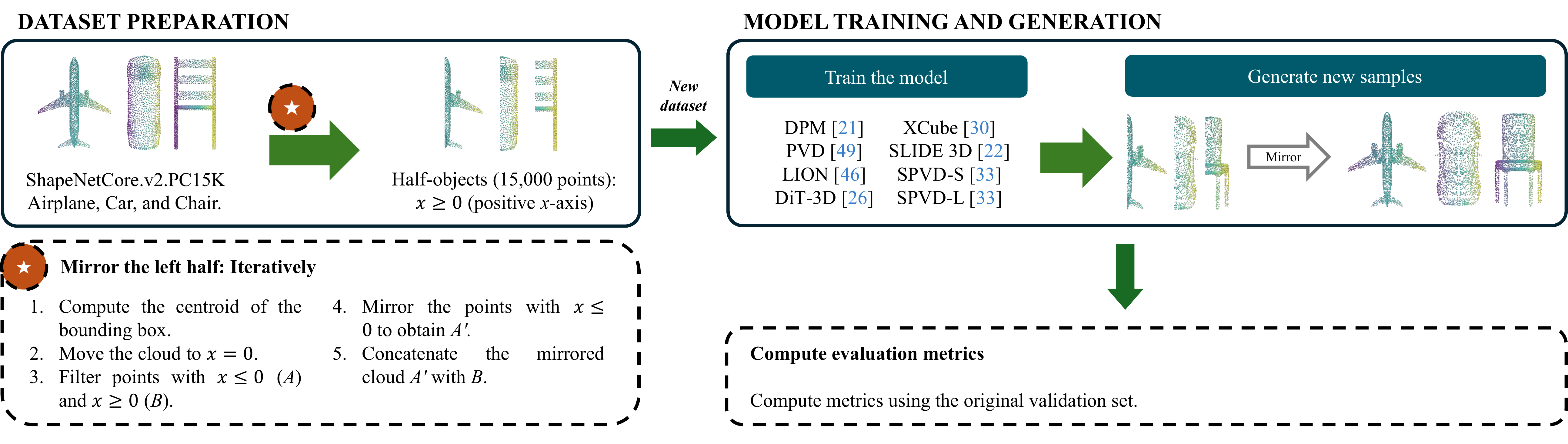}
    \caption{\textbf{Overview of our proposed pipeline.} The process begins with dataset preparation, where objects from ShapeNet are aligned and mirrored along the plane $x = 0$ to obtain right-side half-objects ($x >= 0$), each containing 15,000 points. During generation, the synthesized half-objects are mirrored across the plane $x=0$ to reconstruct full shapes, which are subsequently evaluated using the original validation set.}
    \label{fig:pipeline}
  \end{minipage}
\end{figure*}

For models that first generate a latent representation from an Encoder ($\mathbf{E}$) and then reconstruct shapes via denoising, we examine symmetry at the representation level. Given an input object $\mathbf{X}$ and its reflected counterpart $\mathcal{R}(\mathbf{X})$, we encode both into the latent space ($z=\mathbf{E}(\mathbf{X})$ and $z'=\mathbf{E}(\mathcal{R}(\mathbf{X}))$) and compute a normalized Euclidean distance between their latent codes $||z \, - \, z'||^2_2$. The normalization factor is estimated from random latent pair distances. We can formalize this process as follows:

\begin{equation}
\begin{split}
&d_{L_2}(\mathbf{X}) = {||z \, - \, z'||^2_2}, \\
&\sigma = \mathbb{E}_{z_i,z_j} ||z_i \, - \, z_i'||^2_2, \\
&\mu_{latents} = \frac{1}{N} \sum_{i=1}^{N} \frac{d_{L_2}(X_i)}{\sigma},
\end{split}
\end{equation}

where $\sigma$ is the normalization factor, $\mu_{latents}$ is the average $L_2$ reported in \tabref{tab:symmetry-breaking-inspection}, and $N$ corresponds to the number of objects in the validation split for each class. A large normalized latent distance indicates that the encoder does not preserve symmetric consistency, even when the two inputs differ only by a symmetry transformation.

We repeat the same process for the models trained on the mirrored-objects dataset. 

As reported in \tabref{tab:symmetry-breaking-inspection}, training on mirrored objects reduces the latent discrepancy in all three classes, but does not drive it to zero. This means that the encoder becomes more structure-aware, yet still does not treat a shape and its mirrored counterpart as near-equivalent latent states. The symmetry gap is therefore visible not only in final samples, but also in the organization of the latent space.

Together, the experiments reveal that symmetry breaking is not merely a property of the final generated point cloud. It is already reflected in the dynamics of sampling and in the geometry of learned representations. This is the central diagnostic insight of our work.

\section{Data-Centric Symmetry-Based Induction}\label{sec:proposal} 

As shown in \figref{fig:generated_shapes}, objects in the ShapeNet dataset exhibit clear reflection symmetry with respect to the plane $x=0$. However, as showcased in the previous sections, current 3D generative models fail to preserve this structural property after training or sampling. \figref{fig:generated_shapes} also illustrates that the generated shapes (\emph{top}) are often visually inconsistent, with noticeable deformations and missing regions, and lack the symmetry observed in the training data.

Building on this assumption, we hypothesize that training a 3D generative model exclusively on half-objects is sufficient to learn a rich and diverse distribution of partial geometries, which, when reflected across the symmetry plane, yield complete shapes that are both visually plausible and geometrically symmetric. \figref{fig:pipeline} illustrates the overall pipeline of our data-centric symmetry induction approach.

\subsection{Half-Objects Dataset Preparation}\label{sec:data_prepararion} 

Let $\mathcal{X} \subset \mathbb{R}^{3 \times N}$ be the training dataset composed of point clouds with $N = 15,000$ points, drawn from ShapeNet, and let $\mathcal{X}_c \subset \mathcal{X}$ be the subset of point clouds corresponding to that given class $c \in \{airplane, \, car,\, chair\}$. For each point cloud $X \in \mathcal{X}_c$ we filter both its left half (\emph{A}) and its right half (\emph{B}) with respect to the symmetry plane as follows: $A = \{\mathbf{x} \in X \, | \, \mathbf{x} \leq 0\}$, $B = \{\mathbf{x} \in X \, | \, \mathbf{x} \geq 0\}$; where $\mathbf{x}$ denotes the $x$-coordinate of point $\mathbf{x} \in \mathbb{R}^3$. 

We then mirror the left half \emph{A} across the plane $x = 0$ by negating its $x$-coordinates, producing the reflected set \emph{A'}. The resulting half-object $\tilde{X}$ is obtained by concatenating the mirrored points with the original right half: $\tilde{X} = A' \cup B$. We then apply FPS to sample the point cloud back to 15,000 points.




\begin{table*}[t] 
  \centering
  \small
  \caption{Generation results on Airplane, Car, and Chair compared with original models' results using 1-NNA$\downarrow$ and COV$\uparrow$ as metrics. Both CD and EMD are computed as distance metrics, and reported as percentages (\%). Bold values highlight better model-specific results. Red values highlight better class-specific results. Green values highlight competitive results against benchmarks.}
  \label{tab:generation-results}
  \begin{tabular*}{\linewidth}{@{\extracolsep{\fill}} l | cc cc | cc cc | cc cc @{}}
    \toprule
    \multirow{3}{*}{\textbf{Model}} & \multicolumn{4}{c|}{\textbf{Airplane}} & \multicolumn{4}{c|}{\textbf{Car}} & \multicolumn{4}{c}{\textbf{Chair}} \\
    & \multicolumn{2}{c}{1-NNA ($\downarrow$)} & \multicolumn{2}{c|}{COV ($\uparrow$)} & \multicolumn{2}{c}{1-NNA ($\downarrow$)} & \multicolumn{2}{c|}{COV ($\uparrow$)} & \multicolumn{2}{c}{1-NNA ($\downarrow$)} & \multicolumn{2}{c}{COV ($\uparrow$)} \\
    & CD & EMD & CD & EMD & CD & EMD & CD & EMD & CD & EMD & CD & EMD \\
    \midrule
    DPM              & \textbf{87.13} & 88.49 & \textbf{\color{red}44.31} & 30.45 & 77.31 & 69.94 & \textbf{31.50} & 36.42 & 64.84 & 72.92 & 40.97 & 37.52 \\
    HM-DPM           & 86.01 & 80.20 & 43.56 & 38.86 & 75.29 & 70.09 & 32.66 & 36.13 & 63.81 & 71.27 & 44.74 & 40.03 \\
    \midrule
    \emph{S-DPM}     & 90.22 & \textbf{85.02} & 41.58 & \textbf{37.62} & \textbf{70.52} & \textbf{69.65} & 27.17 & \textbf{37.57} & \textbf{64.44} & \textbf{69.78} & \textbf{41.76} & \textbf{41.44} \\
    \midrule
    \midrule
    PVD              & \textbf{79.46} & \textbf{\color{green}71.16} & \textbf{43.32} & \textbf{\color{red}47.77} & \textbf{\color{green}62.72} & \textbf{\color{red}52.46} & 40.75 & \textbf{49.71} & \textbf{\color{green}57.38} & \textbf{\color{red}55.73} & 45.53 & \textbf{48.98} \\
    HM-PVD           & 85.40 & 81.44 & 41.83 & 37.87 & 65.17 & 57.66 & 41.33 & 47.40 & 58.32 & 59.50 & 44.11 & 50.86 \\
    \midrule
    \emph{S-PVD}     & 96.29 & 93.44 & 18.81 & 19.06 & \textbf{\color{green}62.72} & 61.13 & \textbf{41.91} & 45.66 & 57.69 & 60.99 & \textbf{45.84} & 47.10 \\
    \midrule
    \midrule
    LION             & \textbf{\color{red}77.60} & \textbf{\color{red}67.70} & \textbf{\color{green}43.56} & \textbf{\color{green}45.30} & 64.02 & \textbf{55.35} & \textbf{\color{red}43.35} & \textbf{\color{red}50.58} & \textbf{58.87} & \textbf{\color{green}56.67} & 43.80 & \textbf{46.78} \\
    HM-LION          & 85.89 & 83.42 & 36.88 & 35.64 & 69.08 & 64.31 & 36.99 & 43.35 & 65.54 & 65.93 & 39.72 & 42.70 \\
    \midrule
    \emph{S-LION}    & 82.67 & 78.84 & 39.60 & 37.38 & \textbf{63.01} & 61.13 & 38.15 & 42.49 & 62.32 & 60.28 & \textbf{44.27} & 46.47 \\
    \midrule
    \midrule
    DiT-3D           & 99.01 & 99.01 & 9.41 & 10.15 & 97.54 & 96.68 & 10.12 & 19.94 & \textbf{62.87} & \textbf{61.07} & 37.05 & \textbf{44.90} \\
    HM-DiT-3D        & 99.13 & 99.26 & 8.42 & 7.92 & 94.36 & 93.35 & 17.05 & 23.99 & 60.99 & 62.72 & 39.40 & 44.43 \\
    \midrule
    \emph{S-DiT-3D}  & \textbf{83.17} & \textbf{76.49} & \textbf{\color{red}44.31} & \textbf{39.36} & \textbf{63.44} & \textbf{63.58} & \textbf{36.99} & \textbf{41.62} & 62.95 & 64.99 & \textbf{43.17} & 42.86 \\
    \midrule
    \midrule
    XCube            & 84.78 & \textbf{87.50} & \textbf{38.86} & 31.44 & 69.80 & 75.00 & 38.15 & \textbf{42.49} & \textbf{52.28} & \textbf{58.87} & \textbf{48.35} & \textbf{\color{green}48.51} \\
    HM-XCube         & 84.16 & 89.23 & 40.35 & 33.66 & 69.80 & 74.86 & 38.44 & 39.60 & \textbf{\color{red}52.20} & 58.63 & 46.00 & 48.19 \\
    \midrule
    \emph{S-XCube}   & \textbf{\color{green}78.59} & 88.61 & 37.87 & \textbf{33.66} & \textbf{\color{red}56.36} & \textbf{71.97} & \textbf{41.04} & 34.10 & 59.42 & 60.05 & 43.33 & 44.90 \\
    \midrule
    \midrule
    SLIDE 3D         & 91.83 & 93.44 & 39.36 & 29.70 & \textbf{78.61} & \textbf{80.49} & \textbf{38.15} & 33.53 & 59.50 & 65.62 & \textbf{\color{green}47.88} & 45.53 \\
    HM-SLIDE 3D      & 93.44 & 92.57 & 38.86 & 32.67 & 77.17 & 78.32 & 36.13 & 34.10 & 58.08 & 66.48 & 47.88 & 45.37 \\
    \midrule
    \emph{S-SLIDE 3D} & \textbf{90.59} & \textbf{86.88} & \textbf{41.09} & \textbf{34.90} & 79.05 & 80.64 & 36.71 & \textbf{33.82} & \textbf{59.34} & \textbf{62.64} & 46.31 & \textbf{46.31} \\
    \midrule
    \midrule
    SPVD-S           & 100.00 & 99.88 & 7.18 & 15.84 & \textbf{63.58} & \textbf{\color{green}52.60} & \textbf{\color{green}42.20} & \textbf{\color{green}50.00} & \textbf{58.16} & \textbf{59.03} & \textbf{46.94} & \textbf{47.25} \\
    HM-SPVD-S        & 100.00 & 99.63 & 7.18 & 21.78 & 64.02 & 56.07 & \textbf{\color{red}43.35} & 49.42 & 56.83 & 59.42 & \textbf{\color{red}49.00} & \textbf{\color{red}51.02} \\
    \midrule
    \emph{S-SPVD-S}  & \textbf{92.57} & \textbf{82.67} & \textbf{36.14} & \textbf{37.87} & 68.93 & 67.05 & 37.28 & 38.15 & 73.31 & 70.33 & 38.78 & 42.70 \\
    \midrule
    \midrule
    SPVD-L           & 99.75 & 99.63 & 6.19 & 13.86 & 98.55 & 99.13 & 4.05 & 3.18 & \textbf{67.90} & \textbf{65.93} & 39.25 & \textbf{45.37} \\
    HM-SPVD-L        & 100.00 & 99.75 & 6.43 & 12.13 & 98.99 & 99.28 & 4.62 & 2.89 & 95.68 & 96.62 & 7.22 & 8.32 \\
    \midrule
    \emph{S-SPVD-L}  & \textbf{93.56} & \textbf{87.00} & \textbf{30.20} & \textbf{27.72} & \textbf{80.78} & \textbf{71.97} & \textbf{36.13} & \textbf{30.92} & 69.54 & 69.15 & \textbf{39.40} & 39.87 \\
    \bottomrule
\end{tabular*}
\end{table*}

\subsection{Sampling}\label{sec:generation} 

We generated 1,000 samples per class.

\begin{algorithm}
\caption{Point Cloud Generation}
\label{alg_pcsymmetrization}
\begin{algorithmic}[1]
\REQUIRE Input directory $D_{in}$, output directory $D_{out}$, number of points $M=2048$, denormalization parameters $\mu, \sigma$, integer $k=32$
\ENSURE Symmetric point clouds

\STATE Obtain file list $\mathcal{F}$ from $D_{in}$ with extension \texttt{.npy}

\FOR{each file $f \in \mathcal{F}$}

    \STATE Load point cloud $P \in \mathbb{R}^{N \times 3}$

    \STATE \COMMENT{Move the point cloud to the positive half-space}
    \STATE $P_x \gets P_x - \min(P_x)$

    \STATE \COMMENT{Compute Dynamic Shift}
    \STATE $shift \gets \textsc{DynamicShift}(P, k)$
    \STATE $P_x \gets P_x - shift$

    \STATE \COMMENT{Reflect across the plane $x=0$}
    \STATE $P_{mirror} \gets P$
    \STATE $P_{mirror,x} \gets - P_{mirror,x}$

    \STATE \COMMENT{Concatenate original and mirrored point clouds}
    \STATE $P_{full} \gets \text{Concatenate}(P, P_{mirror})$

    \STATE \COMMENT{Resample to 2048 points}
    \STATE $P_{sampled} \gets \text{FPS}(P_{full}, M)$

    \STATE Save $P_{sampled}$ to $D_{out}$

\ENDFOR

\RETURN Set of symmetrized complete point clouds
\end{algorithmic}
\end{algorithm}

\begin{algorithm}
\caption{DynamicShift}
\label{alg_dynamicshift}
\begin{algorithmic}[1]
\REQUIRE Point cloud $P \in \mathbb{R}^{1 \times N \times 3}$, integer $k$
\ENSURE Shift value $\delta$

\STATE $x \gets$ x-coordinates of $P$
\STATE $k \gets min(k, N)$
\STATE Select the $k$ smallest values of $x$
\STATE $\delta \gets$ mean of selected values
\RETURN $\delta$
\end{algorithmic}
\end{algorithm}

To reconstruct complete shapes, we apply the procedure described in \algref{alg_pcsymmetrization}. This procedure mirrors the generated half-object across the symmetry plane $x=0$, effectively inverting the transformation applied during dataset preparation (\secref{sec:data_prepararion}). To prevent geometric collapse near the symmetry boundary and ensure structural completeness, we introduce a \emph{dynamic shift} mechanism that slightly translates the half-object before reflection. This shift is computed as a scaled average of the smallest $k$ x-coordinates, providing a data-adaptive alignment (as presented in \algref{alg_dynamicshift}). After concatenation, we apply FPS to obtain a set of 2,048 points. 

For evaluation, we use the corresponding validation split.


\section{Experimental Results}\label{sec:experiments}

\subsection{Standard Benchmark Metrics}\label{sec:results}

From \tabref{tab:generation-results}, we can observe that the \textbf{S}ymmetry-aware variants (`S-') remain competitive with the original baselines across all three classes under the standard 1-NNA and COV metrics. In several cases, these variants achieve comparable or even better scores, indicating that the proposed intervention improves reflection symmetry preservation without substantially degrading distributional quality or diversity. Moreover, visual inspection suggests that the generated samples are more structurally coherent and exhibit improved reflection symmetry, as illustrated in \figref{fig:generated_shapes} and \figref{fig:ablation_generated_shapes}. However, since 1-NNA and COV primarily evaluate distributional similarity and sample diversity, and do not explicitly capture structural or geometrical properties \cite{zhu_SeaLionSemanticPartAware_2025}, we further apply our symmetry measurement protocol to quantify the degree of reflection symmetry in the generated samples.

\begin{figure*}[h]
    \centering
    \begin{subfigure}{0.325\textwidth}
        \centering
        \includegraphics[width=\linewidth]{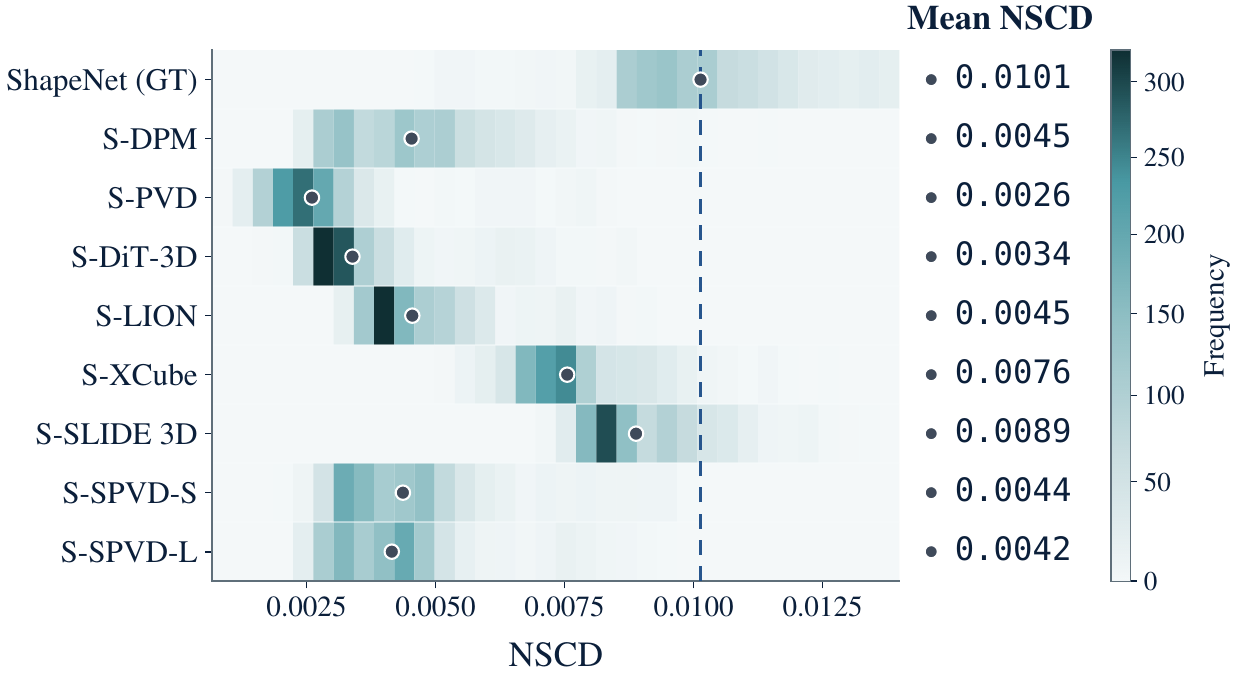}
        \caption{Airplane class}
        \label{fig:s-hist_airplane}
    \end{subfigure}
    \hfill
    \begin{subfigure}{0.325\textwidth}
        \centering
        \includegraphics[width=\linewidth]{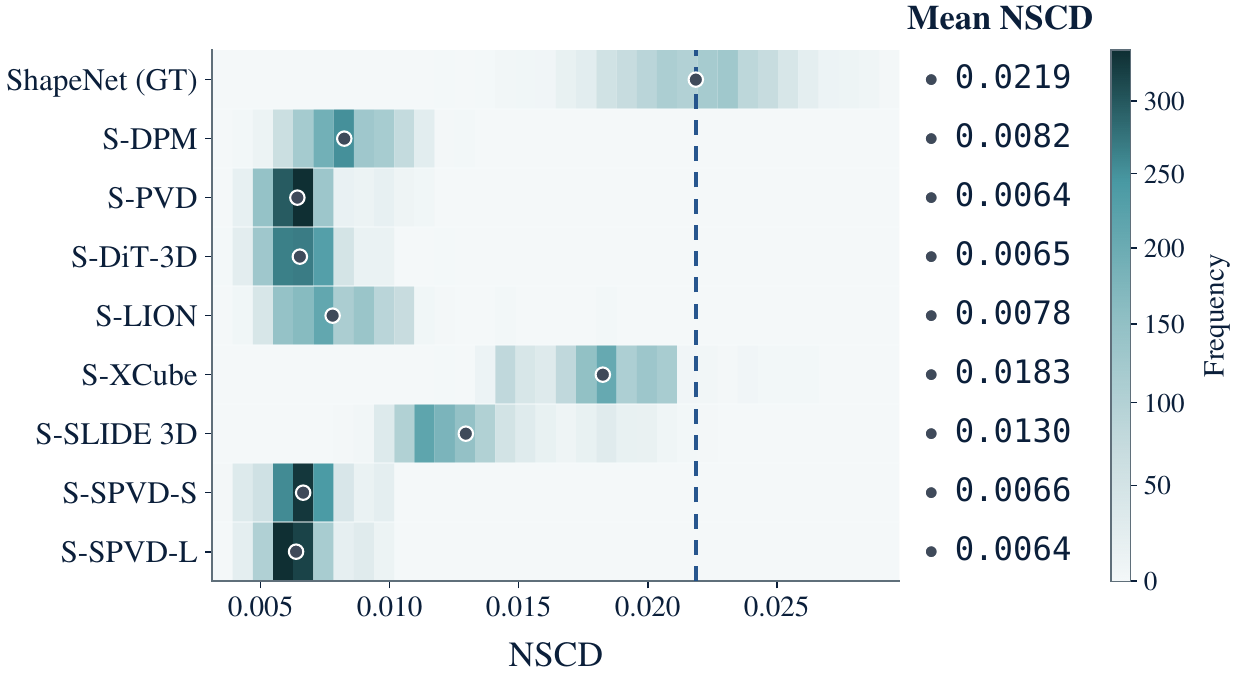}
        \caption{Car class}
        \label{fig:s-hist_car}
    \end{subfigure}
    \hfill
    \begin{subfigure}{0.325\textwidth}
        \centering
        \includegraphics[width=\linewidth]{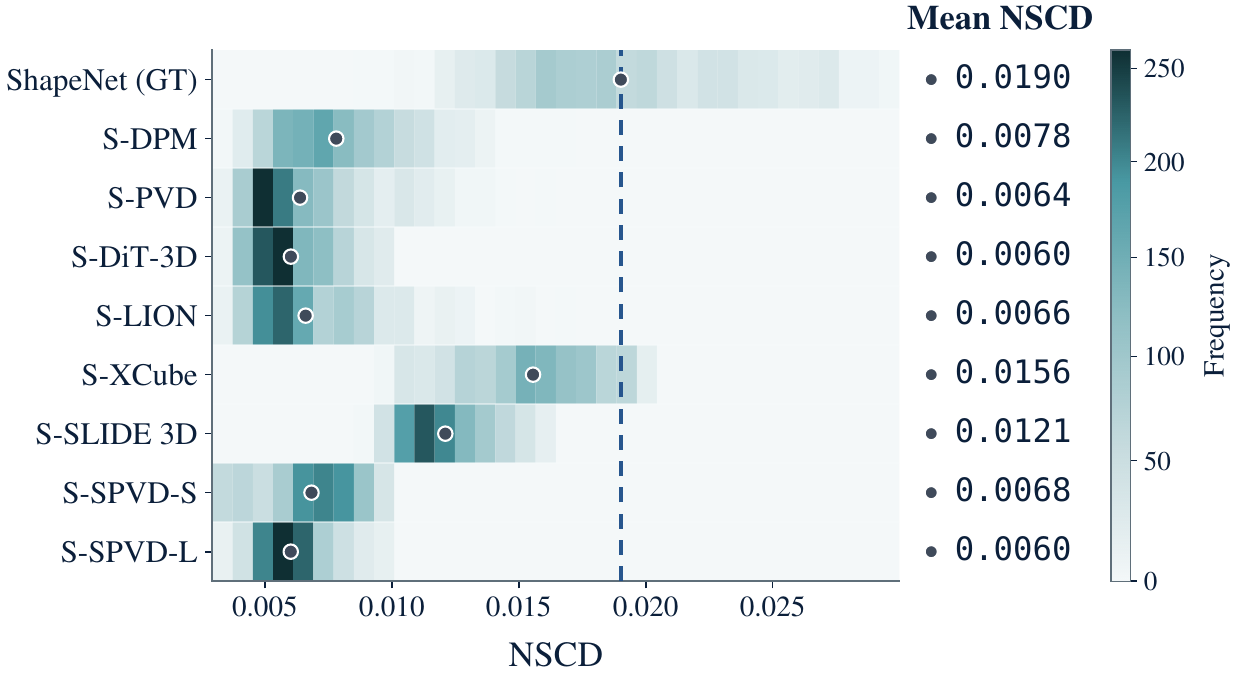}
        \caption{Chair class}
        \label{fig:s-hist_chair}
    \end{subfigure}
    
    \caption{\textbf{Symmetry evaluation across 3D generative models}. Results correspond to the models trained over the \textbf{half-objects ShapeNet dataset}. Ground-truth results from ShapeNet are provided for reference. The vertical dashed line highlights the mean symmetry score of the ground-truth distribution.}
    \label{fig:s-_models_histograms}
\end{figure*}

Distributions in \figref{fig:s-_models_histograms} show that symmetry-based variants generate significantly more symmetric shapes across all three categories, shifting leftward, even from the ground-truth distributions. This indicates a substantial reduction in the mean NSCD values and, consequently, improved reflection symmetry. Importantly, these improved symmetry scores should not be interpreted as evidence that the models internally learn reflection symmetry; rather, symmetry is induced by reflection-based reconstruction of the learned half-shape during generation.

Notably, the symmetry-based variants often achieve mean NSCD values lower than those of the original ShapeNet reference distribution. This is expected to some extent because the final shapes are explicitly reconstructed by reflection. Nevertheless, the behavior of XCube and SLIDE 3D is particularly informative. As discussed in \secref{sec:symmetry_current_3dgenmodels}, these models already produced relatively low NSCD values under the original training setup, possibly due to architectural or training biases that favor smoother and more regular shapes. After training on the half-object dataset, their mean NSCD values remain closer to or below the ground-truth reference, suggesting that the proposed intervention further strengthens reflection consistency. However, this does not necessarily imply that these models explicitly internalize symmetry as a prior; rather, it indicates that the half-object training distribution provides a stronger and more direct symmetry-inducing mechanism.

Overall, these results support the main claim of our work: standard generative benchmarks and symmetry-aware evaluation capture different aspects of model behavior. While 1-NNA and COV assess distributional similarity and diversity, the proposed symmetry measurement protocol reveals whether generated objects preserve high-level geometric structure. Therefore, both evaluation protocols should be interpreted jointly rather than in isolation.





\subsection{Ablation Study on Half-Mirrored Objects from 3D Generative Models}

\begin{figure*} 
    \centering
    \includegraphics[width=\linewidth]{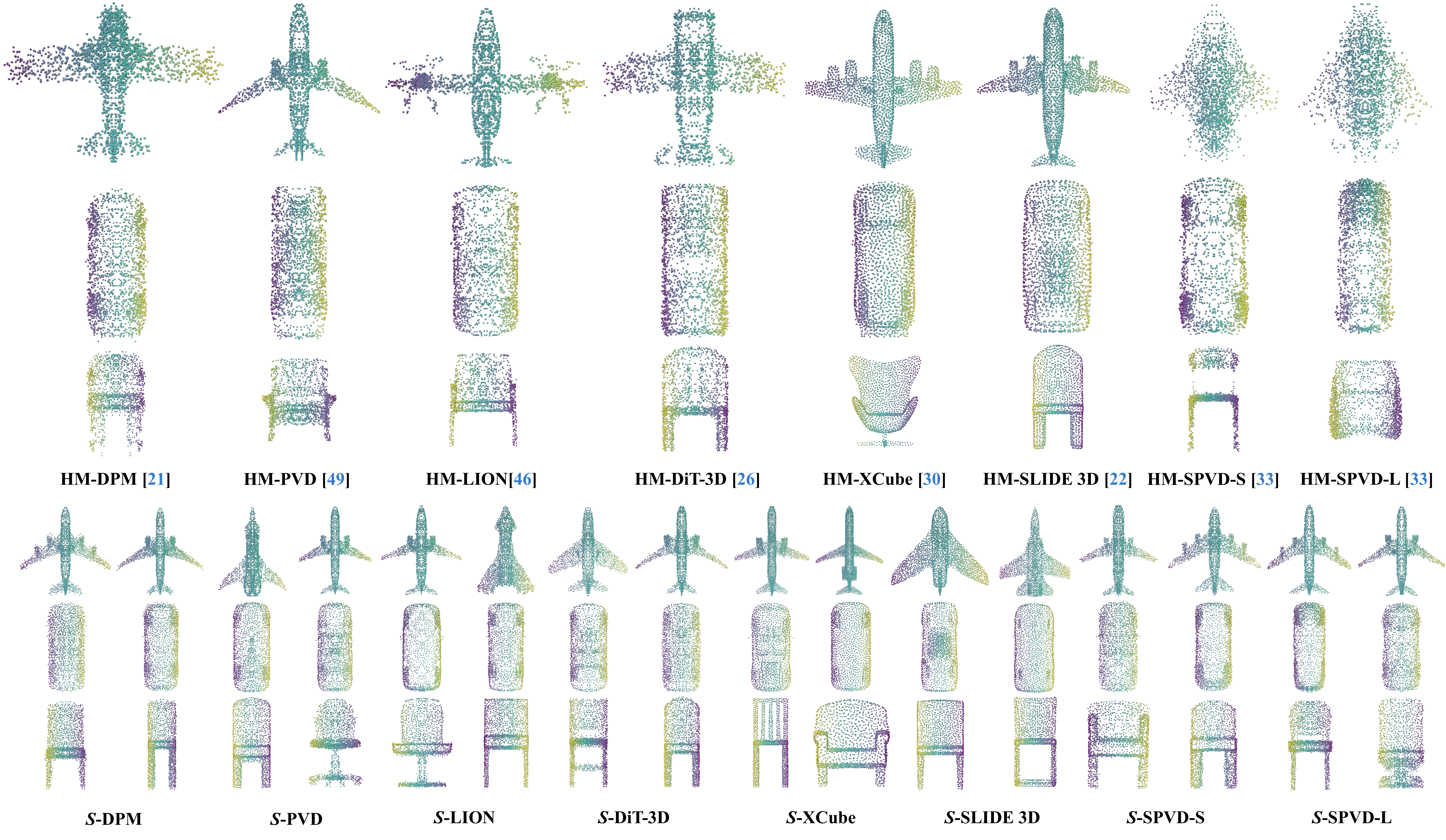}
    \caption{Qualitative comparison between post-hoc half-mirroring and our data-centric symmetry-based approach. \emph{Top}: Half-mirrored reconstructions obtained from samples generated by the original models. \emph{Bottom}: Objects generated by our proposed approach.}
    \label{fig:ablation_generated_shapes}
\end{figure*}

At this point, a natural question arises: \emph{what if we split the generated objects analyzed in \secref{sec:symmetry_current_3dgenmodels} into two halves, retain the half containing the greater number of points, and then reflect it to reconstruct a full object?}

To answer this question, we conduct an ablation experiment across all models and the three ShapeNet categories. We deliberately retain the half with the greater number of points because, when splitting the generated objects along the symmetry plane, we observe that one side often contains considerably more points than the other. This imbalance is itself an important diagnostic signal; if 3D generative models were symmetry-aware, the number of points on both sides of the symmetry plane should remain approximately balanced.

We then resample the reconstructed objects to 2,048 points and compute the metrics reported in \tabref{tab:generation-results}. The results reveal an interesting behavior. In some cases, the \textbf{H}alf-\textbf{M}irrored variants (`HM-') achieve scores comparable to those of the original generated objects, but they also degrade performance in others, indicating that post-hoc symmetrization (splitting and mirroring) does not provide a consistently reliable solution. Furthermore, a qualitative inspection of the generated shapes, as depicted in \figref{fig:ablation_generated_shapes} (\emph{top}), reveals a critical limitation; for many models across the three categories, local geometric errors are duplicated on the reflected side. In other words, if the retained half contains missing parts, incomplete or deformed regions, or outlier points, these artifacts are mirrored and reproduced in the reconstructed object. However, some models, such as XCube and SLIDE 3D in the Airplane and Chair classes, already exhibit good results when symmetrizing the outputs.

This highlights the relevance of our data-centric symmetry-based approach, in which the models learn a rich and diverse distribution of partial geometries rather than relying on symmetry as a post-processing step. Additional results from our approach are included for comparison (\emph{bottom}). 

The results also emphasize the limitations of standard benchmark metrics. Although these metrics are useful for evaluating distributional similarity and diversity, they primarily operate at the level of point-set distances and may fail to capture higher-level structural properties such as symmetry, geometric continuity, and part consistency. As a result, a model can obtain competitive quantitative scores while still producing shapes with evident structural artifacts.

\section{Discussion}\label{sec:discussion}

The main empirical result of this study is that reflection symmetry behaves as a strong prior in ShapeNet, yet is not reliably preserved by current 3D generative models. This observation is consistent across output-space auditing, controlled-data verification, and training-dynamics analysis. This indicates that the symmetry gap is not a marginal artifact of one benchmark or architecture; it is a recurring structural failure mode. This matters because geometry consistency is not merely an aesthetic preference. In many 3D applications, it is tied to shape plausibility, part coherence, and the physical or semantic integrity of generated objects.

The controlled mirrored-dataset experiments reveal that stronger symmetry in the training distribution does not automatically induce symmetry-preserving generation. If the gap were simply inherited from imperfect data, explicitly symmetrizing the dataset should have largely removed or reduced it. On the same line, the mechanism-inspired diagnostic results provide a complementary view of this failure, suggesting that symmetry is not stably organized in the representation space either. From a broader perspective, these diagnostics help shift the analysis from whether the model fails to preserve symmetry toward how this failure becomes visible across different stages of the generative pipeline.

The data-centric half-object intervention is effective precisely because it does not require any modifications to the model architecture or training objective. The gains in NSCD across backbones show that this strategy is simple, yet effective and practically useful. At the same time, improvements in metrics highlight its potential. Moreover, this approach aligns with recent literature advocating for inductive biases in deep generative frameworks \cite{allingham_GenerativeModelSymmetry_2024, lu_StructurePreservingDiffusion_2025}. Further research in this area should consider symmetry as a fundamental design principle for network architectures \cite{diop_GeometricGenerativeModels_2024}. Our data-centric approach lays the foundations for an alternative, symmetry-based design strategy that is simple yet effective. 
Overall, its gains should be interpreted jointly with standard evaluation metrics; since stronger symmetry may induce a trade-off with diversity and fidelity to the original data distribution as reported in \tabref{tab:generation-results}.

Finally, although this work focuses on planar bilateral reflection symmetry, 3D objects may exhibit richer symmetry structures. Therefore, the proposed NSCD score should not be interpreted or used as a symmetry detector or pose-invariant metric; instead, it is used under aligned ShapeNet conditions. Extending this evaluation to non-planar symmetries would require broader transformation searches, object-specific symmetry estimation, or category-dependent canonicalization strategies \cite{podolak_PlanarreflectiveSymmetryTransform_2006, mitra_Symmetry3DGeometry_2013, niu_SymmetryawareAlignmentMethod_2023, zhou_NeRDNeural3D_2021, je_RobustSymmetryDetection_2024}. We leave the integration of these broader symmetry notions into 3D generative evaluation and training for future work.



\section{Conclusion}\label{sec:conclusion}

In this work, we examined whether current 3D generative models preserve structural priors already present in the training data, such as reflection symmetry. We introduced a symmetry measurement protocol based on the Chamfer distance and showed that several representative backbones exhibit a clear symmetry gap relative to the training distribution. By conducting experiments on a controlled mirrored-objects dataset, examining training dynamics, and using mechanism-inspired diagnostic tests at the sampling and latent-representation levels, we show that this gap is not solely explained by data imperfections but can also be induced by the learned generative process itself.

We further proposed a data-centric symmetry-based intervention based on half-object training to alleviate this gap, reconstructing complete shapes through reflection. This simple yet effective strategy substantially improves geometric consistency and visual plausibility while remaining competitive under standard metrics. Our results support the claim that symmetry should be treated as an explicit axis of evaluation for 3D generation, rather than relying solely on diversity and distribution metrics.

\paragraph{\textbf{Limitations and future directions.}} While our work excels over baselines in terms of generation quality and reflection symmetry, it also has limitations. First, the approach relies on reflection symmetry and the assumption that objects are symmetric with respect to a known plane, typically $x=0$, which may not generalize to categories with weak or complex symmetry. As discussed in previous sections, objects with alternative dominant symmetry planes could be canonicalized before evaluation through category-specific symmetry-plane estimation, followed by dataset preprocessing and model retraining under the new canonical alignment. Second, a further direction is to compare our data-centric intervention with alternative symmetry-inducing strategies, including symmetry-aware losses, direct post-processing symmetrization, and architectural mechanisms that explicitly preserve equivariance or reflection consistency. Third, although our study is exhaustive and broad enough to show that current 3D generative models are not symmetry-aware, our mechanism-inspired diagnostics remain limited in scope. Specifically, we do not perform mechanistic-interpretability analyses in the strict sense of the term, such as activation patching, causal ablations, sparse feature decomposition, or circuit tracing. Finally, future work should extend the analysis of commonly used evaluation metrics to identify why they may fail to capture structural properties such as symmetry, geometric continuity, and part consistency.

\paragraph{\textbf{Broader impact.}} This paper is the start of an important conversation about symmetry preservation in current 3D generative models and the use of mechanism-inspired diagnostics to complement output-space evaluation. Despite the main limitations of working with known symmetry planes, there is clearly much more to be done. All that we hope is that our work will start a dialog about this very important and underappreciated prior.



\section*{Acknowledgements}

This work was supported by ANID Chile - Fondecyt Regular N° 1251263, the National Center for Artificial Intelligence CENIA FB210017, Basal ANID, and the Shape Vision Lab from the University of Chile. Nicolas Caytuiro acknowledges the National Doctoral Scholarship Program of the Chilean National Agency for Research and Development (ANID).

This work was supported by the Patagón supercomputer of Universidad Austral de Chile (FONDEQUIP EQM180042).

\section*{Declaration of competing interests}

The authors declare no competing interests.

\section*{Declaration of Generative AI and AI-assisted technologies in the manuscript preparation process}

During the preparation of this work, the authors used ChatGPT and Gemini in order to improve grammar and language readability. After using these tools, the authors reviewed and edited the content as needed and take full responsibility for the content of the published article.

\section*{Data availability}

Exploratory study results, code, and data will be made publicly available upon acceptance of the article.

\appendix

\section{Supplementary data}
The following is the Supplementary material related to this article.

\printcredits

\bibliographystyle{cas-model2-names}

\bibliography{cas-refs}



\end{document}